\documentclass[11pt]{article}
\usepackage[final]{acl}
\usepackage{times,latexsym}
\usepackage[T1]{fontenc}
\usepackage[utf8]{inputenc}

\usepackage{amsmath,amsfonts,bm}

\def\eqref#1{equation~\ref{#1}}

\def\1{\bm{1}}

\DeclareMathAlphabet{\mathsfit}{\encodingdefault}{\sfdefault}{m}{sl}
\SetMathAlphabet{\mathsfit}{bold}{\encodingdefault}{\sfdefault}{bx}{n}

\usepackage{amsmath,amssymb,mathtools}
\usepackage{float}
\usepackage{graphicx}
\usepackage{multirow}
\usepackage{algorithm}
\usepackage[noend]{algpseudocode}
\usepackage{microtype}
\usepackage{array}
\usepackage{longtable}
\usepackage{booktabs}
\usepackage{pifont}
\usepackage{makecell}

\newcommand{\sg}{\operatorname{sg}}
\newcommand{\ours}{LEAF}
\definecolor{cmarkgreen}{HTML}{1B9E4B}
\definecolor{xmarkred}{HTML}{C00000}
\newcommand{\cmark}{\textcolor{cmarkgreen}{\ding{51}}}
\newcommand{\xmark}{\textcolor{xmarkred}{\ding{55}}}
\newcommand{\tail}{\operatorname{tail}}

\title{Explore with Memory, Learn without It: In-Context Exploration and Self-Distillation for LLM Reasoning}

\author{Yutong Chen$^1$ \quad Zhichao Xu$^2$ \quad Rui Liu$^1$ \quad Kunpeng Liu$^1$\thanks{Corresponding author.} \\
$^1$Department of Computer Science, Clemson University\\
$^2$Kahlert School of Computing, University of Utah\\
\texttt{\{yutong2, rliu3, kunpenl\}@clemson.edu} \quad \texttt{zhichao.xu@utah.edu}
}

\begin{document}
\maketitle

\begin{abstract}
Exploration bounds learning efficiency in reinforcement learning with verifiable rewards (RLVR).
Current methods explore mostly through independent sampling, without fully using the information gathered along the way, such as known wrong answers and environment feedback.
We propose \ours{}, which injects a dynamic, problem-specific \emph{memory} into the context before every rollout.
The memory contains the most recent response, its reward, and other environment feedback. It is updated after every verifier call.
We decouple policy updates from rollouts and update the policy through self-distillation.
On four benchmarks and two open models, \ours{} accelerates exploration by about $4\times$ compared with a random sampling baseline. On unseen test sets, it improves accuracy over the strongest baseline by 1.59 and 1.83 percentage points on the two models, while using less compute than any RL baseline.
These results highlight the potential of \ours{} in improving both exploration and learning efficiency.

Code is available at: \url{https://github.com/on1262/LEAF}.

\end{abstract}

\section{Introduction}

Reinforcement learning with verifiable rewards (RLVR) has become the main recipe for improving the reasoning and coding abilities of large language models (LLMs)~\citep{shao2024deepseekmath,guo2025deepseekr1,yu2025dapo,zheng2025gspo,gao2025sapo}.
It maximizes the expected reward, which must be estimated at each rollout step. Since this expectation cannot be computed exactly, most RLVR methods estimate it with Monte Carlo methods~\citep{Metropolis01091949}, which require sampling several responses independently for each prompt.

In general RL, estimating advantages from rewards is challenging. Group-based methods compute advantages using a group of independent responses to the same prompt~\citep{shao2024deepseekmath,yu2025dapo,zheng2025gspo}. A larger group yields more accurate estimates but reduces rollout efficiency. Single-rollout methods such as PPO use a value function~\citep{schulman2017}. However, this requires a critic network. Learning an accurate critic is also challenging when training LLMs~\citep{shao2024deepseekmath,kazemnejad25a}. 

Research on iterative refinement investigates how to improve later responses using earlier attempts. \citet{kumar2024score} and \citet{ekbote2025murphy} use multi-turn reinforcement learning to train revision behavior, including intrinsic self-correction and refinement guided by execution feedback. Some methods learn from feedback and transfer knowledge into parameters. \citet{song2026rltf} learns from feedback through self-distillation or feedback modeling. \citet{shi2026erl} uses reinforcement learning combined with self-distillation. 

Self-distillation methods~\citep{hubotter2026sdpo,zhao2026opsd,he2026sdzero,liu2026pmd,wang2026skillsd,jiang2026skald,wu2026seed} use the same model with privileged context, such as feedback, memories, or skills, as the teacher for an unconditioned student. Although self-distillation does not require duplicated responses, model behavior changes only during the policy update phase, without fully leveraging the model's in-context learning ability.
Methods such as those proposed by \citet{dilkes2026ice} and \citet{liu2026empo} change exploration behavior using a fixed instruction or a self-generated tip. However, they still use duplicated group sampling. As Fig.~\ref{fig:compare} shows, our method is the only one of these designs that combines rich feedback, inference adaptation and a single rollout per step.

\begin{figure}[h]
\centering
\footnotesize
\setlength{\tabcolsep}{4pt}
\begin{tabular}{@{}lccccc@{}}
\toprule
 & \makecell{Random\\sampling} & SDPO & SCoRe & RLVR & Ours \\
\midrule
Rich feedback       & \xmark & \cmark & \xmark & \xmark & \cmark \\
\makecell[l]{Inference\\adaptation} & \xmark & \xmark & \cmark & \xmark & \cmark \\
Single rollout      & \cmark & \xmark & \xmark & \xmark & \cmark \\
\bottomrule
\end{tabular}
\caption{Properties of exploration methods as evaluated here. \emph{Rich feedback}: uses textual feedback such as test output, not only the scalar reward. \emph{Inference adaptation}: the rollout context changes with earlier attempts on the same problem, without a parameter update. \emph{Single rollout}: one verified answer per problem per step.}
\label{fig:compare}
\end{figure}

In our experiments, independent sampling tends to fail with the same wrong answer.
For example, with random sampling on MBPP+, 77.3\% (Qwen3-8B) and 78.0\% (Olmo-3-7B-Instruct) of the failed samples reproduce exactly the failing test output of an earlier sample for the same problem, and on MATH-500 47\% of the wrong answers repeat an earlier wrong answer.
When the previous answer is in the context, adding its feedback significantly reduces the rate of repeating that wrong answer, from 62.8\% to 35.9\% on average over MATH-500, LCB and MBPP+ and the two models. Refer to Appendix~\ref{app:repeat} for details.

Motivated by the strong effect of in-context learning, this work asks whether \textbf{exploration can be improved by changing the \emph{context} of each rollout}.
As illustrated by Fig.~\ref{fig:method}, our method keeps a memory for each problem that contains the last response, its reward and its environment feedback. 
During rollout, each problem only generates one sample per step. We update the memory based on previous attempts on the same problem.
The memory is written into the rollout context as a short dialogue, so each new attempt revises the previous one, as the in-context refinement~\citep{shinn2023reflexion,madaan2023selfrefine}.
During the policy update, the policy conditioned on the memory serves as the teacher, and self-distillation~\citep{agarwal2024gkd,snell2022context,zhao2026opsd} moves what the memory provides into the parameters. 
We name it \ours{}, short for \textbf{L}earning from \textbf{E}arlier \textbf{A}ttempts and \textbf{F}eedback, because each new attempt learns from the previous attempt and the feedback on it in context, and the policy update then learns from these attempts in its parameters.

We evaluate \ours{} on four benchmarks, MATH-500 (mathematics), LiveCodeBench and MBPP+ (code generation) and SciKnowEval (science), with Qwen3-8B and Olmo-3-7B-Instruct.
In exploration tasks, \ours{} raises the cumulative pass rate over Random Sampling by 4.88 pp on average and reaches the final pass rate of Random Sampling with four times fewer verifier calls.
In generalization tasks on unseen test sets, \ours{} has the highest overall test accuracy, which is 1.59 pp (Qwen3-8B) and 1.83 pp (Olmo-3-7B-Instruct) above the strongest baseline.
We perform ablation studies on key components such as the feedback, the last response and the restart schedule. Results show that removing any of them lowers the final pass rate on both models.

\section{Method}

\begin{figure}[t]
\centering
\includegraphics[width=\columnwidth]{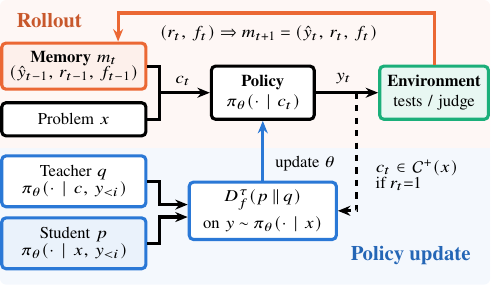}
\caption{Overview of \ours{}. \textbf{Rollout}: each problem keeps a memory of its last response, reward and feedback, which forms the context of the next attempt. The memory is updated after each trial. \textbf{Policy update}: after all rollout steps are finished, the teacher conditioned on a successful context teaches the student without context through self-distillation.}
\label{fig:method}
\end{figure}

As illustrated in Fig.~\ref{fig:method}, in this section we introduce how \ours{} explores and how it updates the parameters. Refer to Appendix~\ref{app:method} for implementation details and the exact prompts.

\subsection{Preliminary}

We first introduce the setting and notation.
Given a set of problems $\mathcal{D}$, each problem receives a budget of $K$ verifier calls, and each call grades one answer.
At step $t$, the environment returns a rollout reward $r_t\in\{0,1\}$ and feedback $f_t$ for the answer $y_t$.
We measure exploration ability by the cumulative pass rate
\begin{equation}
\mathrm{pass}@k=\frac{1}{|\mathcal{D}|}\sum_{x\in\mathcal{D}}\max_{t\le k}s_t(x),
\label{eq:pass}
\end{equation}
where $s_t(x)\in\{0,1\}$ indicates whether $y_t$ is correct.
The environment feedback may expose $s_t$ only partially. In code benchmarks the reward comes from public tests while $s_t$ also requires hidden private tests, so $\mathrm{pass}@k$ is not a simple aggregate of observed rewards.

\subsection{Explore with memory}

For problem $x$ at step $t$, the context $c_t$ is a short dialogue: the problem $x$ as the first user turn, $\hat y_{t-1}$ as the assistant turn, and a final user turn with the verdict $r_{t-1}$, the feedback $f_{t-1}$ and an instruction to check the problem again and give a corrected complete answer. We define the memory as $m_t=(\hat y_{t-1},r_{t-1},f_{t-1})$, which is updated based on previous attempts. In each step, the policy generates $y_t\sim\pi_\theta(\cdot\mid c_t)$, and the verifier call on $y_t$ produces $m_{t+1}$.

To prevent long previous answers from degrading the revision, $\hat y_{t-1}$ in memory is a compressed response based on the raw response $y_{t-1}$. We keep the last $L=2{,}000$ tokens of a longer answer in $\hat y_{t-1}$ because the final result, code or conclusion usually sits at the end. 

When each attempt revises the previous one, the memory can also hold the model on one line of solution.
To keep previous attempts from misleading subsequent ones, we therefore reset the memory every $M=5$ steps. At a restart step, the policy drafts an answer $\tilde y\sim\pi_\theta(\cdot\mid x)$ and revises it once with a blind instruction that asks it to check its own reasoning without any feedback. Only the revision is sent to the verifier.

Exploration with memory can be used alone or together with a policy update. In our experiments, updates during exploration changed the final pass rate little (Section~\ref{sec:interleaved}), so by default \ours{} explores with fixed parameters and updates the policy after all rollouts are done.

\subsection{From memory to parameters}

The memory changes the exploration behavior, but it cannot be applied to a new problem. We distill the experience of rollouts into parameters by self-distillation. Unlike methods that interleave sampling and training, we first finish the rollout phase and then perform the policy update. This keeps the implementation simple and avoids waiting for long outputs inside every update step.

After a rollout of $K$ steps on the training problems, we collect the set $\mathcal{C}^{+}(x)$ of contexts $c_t$ whose following response $y_t$ is correct for each problem $x$. Each context holds the memory that produced a correct answer, i.e. last response and feedback. At each update step, we sample a batch $\mathcal{B}$ of problems.
For each $x\in\mathcal{B}$ we randomly draw a context from $\mathcal{C}^{+}(x)$ and a student answer $y\sim\pi_\theta(\cdot\mid x)$ generated from the plain prompt. Then we distill the token distribution from teacher to student.

At token position $i$ of $y$, the student distribution is $p_i=\pi_\theta(\cdot\mid x,y_{<i})$ and the teacher distribution is $q_i=\sg[\pi_\theta(\cdot\mid c,y_{<i})]$, where $\sg$ stops the gradient. We compare them with a clipped $f$-divergence over the full vocabulary $\mathcal{V}$,
\begin{equation}
D_f^{\tau}(p_i\,\|\,q_i)=\sum_{v\in\mathcal{V}}\min\!\left(q_{i,v}\,f\!\left(\frac{p_{i,v}}{q_{i,v}}\right),\,\tau\right).
\label{eq:fdiv}
\end{equation}
We use $f(u)=u\log u$ by default, which gives the reverse KL divergence, i.e., $\mathrm{KL}(p_i\,\|\,q_i)$.
We also compare other KL losses in Section~\ref{sec:loss}. Following OPSD~\citep{zhao2026opsd}, we cap each summand of Eq.~(\ref{eq:fdiv}) at a threshold $\tau$ to avoid length collapse. Entries above $\tau$ receive no gradient. We use $\tau=10^{-3}$ on MATH and no cap ($\tau=+\infty$) on LiveCodeBench and SciKnowEval. The final loss is described as follows:

\begin{equation}
\mathcal{L}(\theta)=\frac{1}{\sum_{x\in\mathcal{B}}|y^{x}|}\sum_{x\in\mathcal{B}}\sum_{i=1}^{|y^{x}|}w^{x}_{i}\,D_f^{\tau}\big(p^{x}_{i}\,\|\,q^{x}_{i}\big),
\label{eq:loss}
\end{equation}
where $y^{x}$, $p^{x}_i$ and $q^{x}_i$ belong to problem $x$, and $w^{x}_{i}$ is a truncated importance weight (Appendix~\ref{app:method}).
Algorithm~\ref{alg:ours} gives the complete procedure.

\begin{algorithm}[h]
\caption{\ours{}. $c_t=\langle u_1,a,u_2\rangle$: user, assistant and user turns.}
\label{alg:ours}
\footnotesize
\begin{algorithmic}[1]
\Require problems $\mathcal{D}$, policy $\pi_\theta$, $K$, $M$, $L$, $S$
\For{each $x\in\mathcal{D}$ \textbf{in parallel}}
  \State $\mathcal{C}^{+}(x)\gets\emptyset$
  \For{$t=1,\dots,K$}
    \If{$(t-1)\bmod M=0$} \Comment{restart}
      \State $\tilde y\sim\pi_\theta(\cdot\mid x)$
      \State $c_t\gets\langle x,\,\tail_L(\tilde y),\,\text{blind instruction}\rangle$
    \Else
      \State $\hat y_{t-1}\gets\tail_L(y_{t-1})$
      \State $c_t\gets\langle x,\,\hat y_{t-1},\,(r_{t-1},f_{t-1},\text{instruction})\rangle$
    \EndIf
    \State $y_t\sim\pi_\theta(\cdot\mid c_t)$; $(r_t,f_t)\gets\mathrm{Env}(x,y_t)$
    \If{$r_t=1$ \textbf{and} $y_t$ not truncated} $\mathcal{C}^{+}(x)\gets\mathcal{C}^{+}(x)\cup\{c_t\}$ \EndIf
  \EndFor
\EndFor
\For{$j=1,\dots,S$} \Comment{policy update}
  \State sample batch $\mathcal{B}\subseteq\{x:\mathcal{C}^{+}(x)\neq\emptyset\}$
  \State $c\sim\mathrm{Unif}(\mathcal{C}^{+}(x))$, $y\sim\pi_\theta(\cdot\mid x)$ for each $x\in\mathcal{B}$
  \State update $\theta$ with one gradient step on Eq.~(\ref{eq:loss})
\EndFor
\State \Return $\pi_\theta$
\end{algorithmic}
\end{algorithm}

\section{Experiments}

\begin{figure*}[t]
\centering
\includegraphics[width=\textwidth]{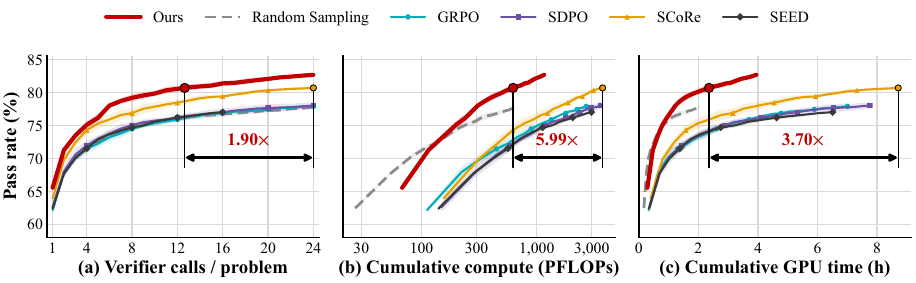}
\caption{Cumulative pass rate on four benchmarks against (a) verifier calls per problem, (b) cumulative compute in PFLOPs (log scale) and (c) cumulative GPU time. Both axes are averaged over the two models at each step. }
\label{fig:explore}
\end{figure*}

We compare our method with baselines on two tasks.
In the exploration task, we evaluate how quickly \ours{} discovers correct solutions. To measure this ability, we compute pass@k after each step.
In the in-distribution generalization task, we investigate whether the knowledge in rollout trajectories can be transferred to model parameters.
The model first explores and updates policy on the training set, then we evaluate its accuracy on a held-out test set.
The results of both tasks (Fig.~\ref{fig:explore}, Table~\ref{tab:explore} and Table~\ref{tab:general}) are averaged over three runs with different random seeds, as is the length analysis in Table~\ref{tab:length}. Other ablation studies use a single seed.

\subsection{Exploration tasks}

\subsubsection{Settings}

\paragraph{Datasets.}
We collect four datasets for the exploration tasks: the 500 problems from MATH-500~\citep{hendrycks2021math,lightman2024verify}, 131 problems from release v6 of LiveCodeBench (LCB)~\citep{jain2025livecodebench} in the split used by SDPO~\citep{hubotter2026sdpo}, the 378 problems from MBPP+~\citep{liu2023evalplus}, and 400 open-ended problems from SciKnowEval~\citep{feng2024sciknoweval}, with 100 problems each from biology, chemistry, materials, and physics. The SciKnowEval problems are randomly sampled, with 25\% at level L5 and 75\% at level L2. We filter out all multiple-choice and binary problems to prevent the model from guessing answers.

\paragraph{Verifiers and feedback.}
For MATH-500, we use only correctness as feedback. For SciKnowEval, we use a base LLM as a judge and provide only correctness as feedback. For coding benchmarks such as LCB and MBPP+, the feedback includes runtime errors, code, and test inputs and outputs. Only failed tests are included in the feedback. We divide all test cases into public and private sets. Rewards and feedback are based only on public tests. The pass rate is computed based on solutions that pass all tests in both the public and private sets.
For a fair comparison, we assign $K=24$ verifier calls per problem and at most 5 GPU hours per benchmark.  We measure the wall time on one H200 GPU per run.

\paragraph{Decoding settings.}
We use Qwen3-8B~\citep{yang2025qwen3} with thinking disabled and Olmo-3-7B-Instruct~\citep{olmo2025olmo3}. We sample from Qwen3-8B with a temperature of 1.0 and top-$p$ of 1.0, and from Olmo-3-7B-Instruct with a temperature of 0.6 and top-$p$ of 0.95.
The maximum response length is 8,192 tokens for all methods, except when evaluating Olmo-3 on LiveCodeBench, where we use 16,384 tokens because we found that Olmo-3 produces longer reasoning sequences on this benchmark.

\paragraph{Baselines.}
We compare our method with five baselines. \textit{Random Sampling} draws one independent sample per problem per step from the plain prompt.
We include four RL baselines:
\textit{GRPO}~\citep{shao2024deepseekmath} is a standard RLVR baseline. We use the higher clipping bound from DAPO~\citep{yu2025dapo}.
\textit{SDPO}~\citep{hubotter2026sdpo} is the most relevant method, as it also uses environment feedback and self-distillation.
\textit{SCoRe}~\citep{kumar2024score} trains on two-turn trajectories in which the second turn revises the first. We include SCoRe as a self-improvement baseline.
\textit{SEED}~\citep{wu2026seed} is a hybrid method that combines SFT and RL.
For all baselines with policy updates, we use LoRA adapters~\citep{hu2022lora} with rank 64 and a constant learning rate of $10^{-6}$.

\paragraph{Our method.}
\ours{} runs only the rollout part of Algorithm~\ref{alg:ours}. We use a restart interval of $M=5$ and a tail length of $L=2{,}000$ tokens. The variant with parameter updates during exploration is studied in Section~\ref{sec:interleaved}. For a fair comparison, we count the rollout resource usage as a part of our method. Refer to Appendix~\ref{app:explore-setup} for details.

\subsubsection{Results}

\begin{table}[h]
\centering
\footnotesize
\setlength{\tabcolsep}{2pt}
\resizebox{\columnwidth}{!}{%
\begin{tabular}{lrrrrrr}
\toprule
 & \multicolumn{3}{c}{Qwen3-8B} & \multicolumn{3}{c}{Olmo-3-7B-Inst.} \\
\cmidrule(lr){2-4}\cmidrule(lr){5-7}
Method & Pass & GPU h & PFLOPs & Pass & GPU h & PFLOPs \\
\midrule
Random Sampling & 77.44 & \textbf{0.66} & \textbf{625} & 78.15 & \textbf{3.42} & \textbf{676} \\
GRPO & 77.78 & 4.43 & 2,494 & 78.00 & 9.58 & 2,885 \\
SDPO & 77.71 & 5.06 & 3,277 & 78.31 & 10.46 & 3,843 \\
SCoRe & \underline{78.65} & 5.09 & 3,058 & \underline{82.73} & 12.33 & 4,401 \\
SEED & 77.96 & 7.87 & 4,580 & 77.47 & 9.43 & 3,796 \\
Ours & \textbf{80.84} & \underline{1.47} & \underline{1,043} & \textbf{84.50} & \underline{6.43} & \underline{1,275} \\
\bottomrule
\end{tabular}}
\caption{Pass rate (\%) averaged over the four benchmarks and summed exploration cost. Refer to Appendix~\ref{app:explore-results} for results by seed and benchmark.}
\label{tab:explore}
\end{table}

\paragraph{\ours{} is more efficient and less costly than the best baseline.}

\begin{table*}[h]
\centering
\small
\setlength{\tabcolsep}{2.8pt}
\begin{tabular}{lrrrrrrrrrrrr}
\toprule
 & \multicolumn{6}{c}{Qwen3-8B} & \multicolumn{6}{c}{Olmo-3-7B-Instruct} \\
\cmidrule(lr){2-7}\cmidrule(lr){8-13}
Method & LCB$\uparrow$ & MATH$\uparrow$ & SciKE$\uparrow$ & Overall$\uparrow$ & GPU h$\downarrow$ & EFLOPs$\downarrow$ & LCB$\uparrow$ & MATH$\uparrow$ & SciKE$\uparrow$ & Overall$\uparrow$ & GPU h$\downarrow$ & EFLOPs$\downarrow$ \\
\midrule
Base & 19.78 & 84.08 & 74.40 & 59.42 & -- & -- & 21.25 & 89.12 & 69.25 & 59.87 & -- & -- \\
GRPO & \textbf{26.02} & \underline{86.53} & 76.75 & \underline{63.10} & 5.52 & 3.71 & 21.12 & \underline{89.33} & 70.83 & 60.43 & 14.06 & 4.52 \\
SDPO & 25.64 & 81.65 & 72.96 & 60.08 & 5.82 & 4.35 & \underline{23.66} & 82.38 & 66.98 & 57.68 & 14.23 & 5.93 \\
SCoRe & 22.71 & 84.32 & 75.17 & 60.73 & 5.57 & 4.17 & 20.48 & 88.95 & 68.38 & 59.27 & 17.71 & 6.33 \\
SEED & \underline{25.89} & 85.58 & \textbf{76.83} & 62.77 & 6.72 & 4.88 & 22.33 & 89.10 & \underline{72.10} & \underline{61.18} & 14.56 & 5.94 \\
SFT & 22.26 & 76.85 & 75.38 & 58.16 & \textbf{3.32} & \textbf{2.44} & 22.77 & 81.33 & 70.52 & 58.21 & \textbf{13.47} & \textbf{3.00} \\
Ours & 25.45 & \textbf{91.85} & \underline{76.77} & \textbf{64.69} & \underline{3.70} & \underline{2.64} & \textbf{24.05} & \textbf{90.62} & \textbf{74.35} & \textbf{63.01} & \underline{14.03} & \underline{3.15} \\
\bottomrule
\end{tabular}
\caption{In-distribution generalization results. Avg@4 (\%) on test sets. LCB: LiveCodeBench v6. MATH: MATH-500. SciKE: SciKnowEval. GPU h and EFLOPs ($10^{18}$ FLOPs) sum the three training runs. For \ours{} and SFT they include the exploration rollout that collects the training data, and for SFT also rewriting and verifying the solutions. Refer to Appendix~\ref{app:general-results} for results by run seed and training cost by stage.}
\label{tab:general}
\end{table*}

We found that our method has higher exploration efficiency and lower cost. In Table~\ref{tab:explore}, \ours{} reached 80.84\% on Qwen3-8B and 84.50\% on Olmo-3-7B, which surpasses the best baseline, SCoRe, by 2.19 pp and 1.77 pp. In Fig.~\ref{fig:explore}, \ours{} uses far less computation or GPU wall time than baselines. It accelerates the exploration by $3.70 \times$ compared with the best baseline. Since \ours{} explores purely by in-context learning, the computation cost is at the same level as random sampling (grey line). But our method continuously improves with more compute.

\subsection{In-distribution generalization tasks}

\subsubsection{Settings}

\paragraph{Datasets.}
We train on three training sets that are disjoint from the test sets. In the training sets, we use all test cases for reward and feedback. There is no overlap between the training sets and the benchmarks in this task.
For LiveCodeBench, the training set contains 478 problems from releases v2 to v5 of LiveCodeBench. 
For MATH, we randomly extract 500 Level 5 problems from the MATH training split~\citep{hendrycks2021math}.
For SciKnowEval, we extract 400 open problems from Biology/Chemistry/Material/Physics at level L5. We use the same test set as the exploration task

\paragraph{Benchmarks.}
The model trained on each training set is tested on the matching benchmark, LCB v6, MATH-500 or SciKnowEval, which together hold the same 1,031 problems as in exploration.
We sample each problem with four responses and report the avg@4 as generalization performance.

\paragraph{SFT baseline.} To test whether the gain comes from the correct answers collected during exploration rather than from self-distillation, we build an SFT baseline from the same rollouts as \ours{}.
For each successful context and its correct answer in our method, the base model rewrites the answer into a self-contained solution without the earlier attempts and the references to feedback.
We keep only the correct rewrites and train the plain prompt on them with the same LoRA, batch size and number of steps as \ours{}.

\paragraph{Other Experimental Settings.}
Training rollouts and test samples use the decoding settings of the exploration experiments. We compare four RL baselines, GRPO, SDPO, SCoRe and SEED, and an SFT baseline trained on the same rollouts as \ours{}. They use the exploration settings except for a unified learning rate of $10^{-5}$ and a budget of 20 verifier calls per training problem. For SEED, we found its compute time is significantly longer than that of other baselines. We lower the budget to 15 for a fair comparison. \ours{} first explores each training set for $K=20$ steps.
It then runs $S=20$ update steps with a batch size of 128. Refer to Appendix~\ref{app:general-setup} for more details.

\subsubsection{Results}

\paragraph{\ours{} achieves better performance on unseen test sets with lower cost.} 
In Table~\ref{tab:general}, we find that \ours{} reaches an overall avg@4 of 64.69\% on Qwen3 and 63.01\% on Olmo-3, exceeding the best baseline, GRPO on Qwen3 and SEED on Olmo-3, by 1.59 pp and 1.83 pp. \ours{} is lower only on LCB (25.45\% versus 26.02\% for GRPO) and SciKnowEval (76.77\% versus 76.83\% for SEED) with Qwen3-8B.
We found that our method uses significantly less compute. \ours{} uses 2.64 and 3.15 EFLOPs, about 1.9 times less than SEED (4.88 and 5.94 EFLOPs). The GPU time of our method on Olmo-3 is close to that of GRPO and SDPO, but our method achieves higher performance.

\paragraph{\ours{} improves accuracy without relying on better rollout data.}
In Table~\ref{tab:general} we find that SFT reaches an overall avg@4 of only 58.16\% and 58.21\%, which is 6.53 pp and 4.80 pp below \ours{}, although its compute is close to that of \ours{}.
SFT raises the accuracy on LCB and SciKnowEval over Base on both models, but its MATH-500 accuracy falls 7.23 pp and 7.79 pp below Base.
This may be because the rewritten solutions are generated in a context that the student never sees, so imitating them pulls the student away from its own distribution, while \ours{} matches the teacher on the student's own answers.
These results suggest that the gain of \ours{} comes from the distillation method and not from the correct answers alone.

\subsection{Ablation studies}

\subsubsection{Effect of restart, last response, feedback and history length during exploration}

We investigate the effect of the key components in our method. There are four ablations on exploration tasks: the last response, the feedback, the restart interval and the number of past rounds in the context. Each ablation removes one component. Refer to Appendix~\ref{app:abl-components} for details.

\begin{figure}[h]
\centering
\includegraphics[width=\columnwidth]{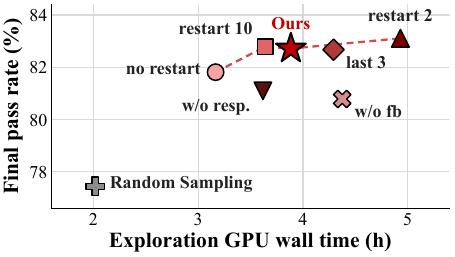}
\caption{Exploration ablations on key components. Final pass rate against exploration GPU time, averaged over the two models. Our method in main experiments is denoted as red star.}
\label{fig:ablation}
\end{figure}

\paragraph{Last response and feedback cause most improvement.}
In Fig.~\ref{fig:ablation} we find that removing the compressed response of the last turn significantly drops performance on both models. We design \ours{} w/o last response as a single user turn with the problem and the feedback without the assistant turn. Its pass rate ends at 79.70\% and 82.49\% instead of 80.94\% and 84.49\% on exploration tasks.
Removing the feedback (\ours{} w/o feedback) lowers the final pass rate more than removing the last response on Qwen3-8B but less on Olmo-3-7B-Instruct, where it lowers the mean accuracy by only 0.46 pp. This may be because the last response lets the model edit a concrete attempt, while feedback alone only says that an unseen attempt has failed.

\paragraph{Restarts cause a trade-off between pass@k and avg@k.}
We find that \ours{} w/o restarts has the highest mean accuracy but a lower final pass rate compared with \ours{}.
Restarting every 2 steps has the highest final pass rate, but it significantly increases GPU wall time from 3.89 to 4.93 hours and drops the mean accuracy by 1.47 pp.
One possible explanation is that a memory never reset keeps refining one line of solution, which helps on solved problems but explores less on the others.
We use restart every 5 steps to keep a balance between pass rate and accuracy.

\paragraph{A longer history does not help consistently.}
We find that keeping the last three assistant and user rounds in the dialogue instead of one lowers the averaged final pass rate (82.71\% to 82.67\%) and raises the GPU time (3.89 to 4.29 hours).
It may be because older rounds repeat attempts that the last round already revised, so they lengthen the context with little new information.
We therefore keep only the last round in our method.

\subsubsection{Effect of response length}

Here we investigate how response length relates to the accuracy gain in exploration rollouts and generalization tasks. In Table~\ref{tab:length}, length and accuracy are compared with a reference (Random Sampling or Base) for each model and benchmark and then averaged. Refer to Appendix~\ref{app:abl-length} for details.

\begin{table}[h]
\centering
\footnotesize
\setlength{\tabcolsep}{4pt}
\begin{tabular}{lrrrr}
\toprule
 & \multicolumn{2}{c}{(a) Exploration} & \multicolumn{2}{c}{(b) Generalization} \\
\cmidrule(lr){2-3}\cmidrule(lr){4-5}
Method & Length & Gain (pp.) & Length & Gain  (pp.) \\
\midrule
GRPO & 1.00 & 0.09 & 1.14 & 2.12 \\
SDPO & 0.99 & 0.21 & 0.98 & $-$0.77 \\
SCoRe & 1.27 & 2.89 & 1.01 & 0.35 \\
SEED & 1.00 & 0.31 & 1.11 & 2.33 \\
SFT & -- & -- & 0.93 & $-$1.46 \\
Ours & 1.17 & \textbf{4.88} & 1.60 & \textbf{4.20} \\
\bottomrule
\end{tabular}
\caption{Response length and accuracy gain over the reference. Length: ratio of the mean response length to that of the reference. Gain: accuracy gain over the reference in pp.}
\label{tab:length}
\end{table}

\paragraph{\ours{} turns extra exploration tokens into the largest pass rate gain.}
In Table~\ref{tab:length}(a), we find that \ours{} writes answers 1.17 times as long as Random Sampling and raises the pass rate by 4.88 pp, which is the largest gain. SCoRe, which also revises an earlier attempt, writes even longer answers (1.27 times) but gains only 2.89 pp. Other baselines such as GRPO, SDPO and SEED keep almost the same length as Random Sampling and gain at most 0.31 pp.

\paragraph{The accuracy gain is related to the length after the policy update} 
In Table~\ref{tab:length}(b), we find that the accuracy gain in generalization tasks is highly related to the response length across methods.
SFT and SDPO shorten the responses and fall below Base by 1.46 and 0.77 pp, SCoRe keeps the length and gains 0.35 pp. GRPO and SEED lengthen the responses by 11 to 14\% and gain 2.12 and 2.33 pp.
\ours{} extends this trend furthest, with responses 1.60 times as long as Base and the highest gain of 4.20 pp.

\subsubsection{Effect of interleaved updates during exploration}
\label{sec:interleaved}

In this section we investigate the effect of updating the parameters during exploration. We update the policy every $N=1,2,5,+\infty$ rollout steps and observe the final pass rate in exploration tasks. Refer to Appendix~\ref{app:abl-interleaved} for details.

\begin{table}[h]
\centering
\footnotesize
\setlength{\tabcolsep}{1.6pt}
\begin{tabular}{lrrrr}
\toprule
 & \multicolumn{2}{c}{Qwen3-8B} & \multicolumn{2}{c}{Olmo-3-7B-Inst.} \\
\cmidrule(lr){2-3}\cmidrule(lr){4-5}
Method & Pass & GPU h & Pass & GPU h \\
\midrule
Random Sampling (ref.) & 77.06 & \textbf{0.64} & 77.81 & \textbf{3.39} \\
Ours (no update) & \textbf{80.94} & \underline{1.47} & \underline{84.49} & \underline{6.30} \\
Ours + update every step & 80.48 & 5.55 & \textbf{85.16} & 13.97 \\
Ours + update every 2 steps & \underline{80.80} & 4.38 & 84.14 & 11.40 \\
Ours + update every 5 steps & 80.53 & 3.40 & 84.25 & 9.70 \\
\bottomrule
\end{tabular}
\caption{Final cumulative pass rate (\%, averaged over the four benchmarks) and exploration GPU hours. }
\label{tab:interleaved}
\end{table}

\paragraph{Updates during exploration add cost but no consistent gain.}
In Table~\ref{tab:interleaved} we find that on Qwen3-8B all update intervals end slightly below no update (80.48\% to 80.80\% versus 80.94\%).
On Olmo-3-7B-Instruct, updates every step end higher (85.16\%) and every 2 or 5 steps lower (84.14\% and 84.25\%) than no update (84.49\%).
The updates raise the GPU time by 2.3 to 3.8 times on Qwen3-8B and by 1.5 to 2.2 times on Olmo-3-7B-Instruct.
These results highlight that updating parameters brings little extra improvement with in-context learning.

\subsubsection{Effect of rollout traces in policy update}
In the policy update, the rollout traces in the context decide how well the teacher answers. This ablation compares the teacher accuracy under three types of traces. Refer to Appendix~\ref{app:abl-traces} for details.
\textit{Successful traces} are the contexts in $\mathcal{C}^{+}(x)$, as in \ours{}.
\textit{Failure traces} are the contexts whose following response is incorrect.
\textit{Random traces} keep the successful traces but replace the feedback with a different one, taken from another step of the same problem or, if none exists, from another problem. On MATH and SciKnowEval the feedback is only the verdict.

\begin{table}[h]
\centering
\footnotesize
\setlength{\tabcolsep}{2.8pt}
\begin{tabular}{lrrrrrr}
\toprule
 & \multicolumn{3}{c}{Qwen3-8B} & \multicolumn{3}{c}{Olmo-3-7B-Inst.} \\
\cmidrule(lr){2-4}\cmidrule(lr){5-7}
Traces & LCB & MATH & SciKE & LCB & MATH & SciKE \\
\midrule
Successful & \textbf{82.98} & \textbf{90.86} & \textbf{83.23} & \textbf{71.03} & \textbf{96.16} & \textbf{67.93} \\
Failure & 39.82 & 60.30 & 67.07 & 44.76 & 79.48 & 50.13 \\
Random & 69.30 & 88.54 & 80.16 & 69.24 & 93.76 & 66.80 \\
\bottomrule
\end{tabular}
\caption{Teacher accuracy (\%) on the training problems under different rollout traces.}
\label{tab:traces}
\end{table}

\paragraph{Memory of successful responses indicates higher teacher accuracy.}
In Table~\ref{tab:traces} we find that successful traces give the highest teacher accuracy on all three training sets and both models, 85.69\% and 78.37\% on average.
Failure traces lower it to 55.73\% and 58.12\%, with the largest drop on LCB with Qwen3-8B (82.98\% to 39.82\%).
Random traces lower it only to 79.33\% and 76.60\%. The largest drop is again on LCB with Qwen3-8B (13.68 pp), where the feedback holds the failed tests. This result reveals that the memory of successful responses leads to higher teacher accuracy in self-distillation.

\begin{table}[h]
\centering
\footnotesize
\setlength{\tabcolsep}{4pt}
\begin{tabular}{lrrrr}
\toprule
 & \multicolumn{2}{c}{Qwen3-8B} & \multicolumn{2}{c}{Olmo-3-7B-Inst.} \\
\cmidrule(lr){2-3}\cmidrule(lr){4-5}
Method & Overall & Trunc. & Overall & Trunc. \\
\midrule
Base & 59.41 & 0.02 & 59.84 & 2.04 \\
Ours (reverse KL) & \textbf{64.92} & 2.25 & \textbf{62.71} & 2.45 \\
Ours w/ forward KL & 63.80 & 1.77 & 60.93 & 2.02 \\
Ours w/ JSD & 59.69 & 0.10 & 62.04 & 1.83 \\
\bottomrule
\end{tabular}
\caption{Ablations on three loss types. Overall test avg@4 (\%) and truncated test responses (\%, mean over the three test sets).}
\label{tab:loss}
\end{table}

\subsubsection{Effect of the divergence in the policy update}
\label{sec:loss}

In this section we investigate the effect of the divergence in Eq.~(\ref{eq:fdiv}) on the policy update. We test 3 settings: Reverse KL, which is used in main experiments, Forward KL and JSD with $\beta=0.5$. All other settings are unchanged. Refer to Appendix~\ref{app:abl-loss} for details.

\paragraph{Reverse KL gives the highest overall accuracy.}
In Table~\ref{tab:loss}, we find that reverse KL reaches 64.92\% and 62.71\%. Forward KL only reaches 63.80\% and 60.93\% and JSD reaches 59.69\% and 62.04\%.
The largest drops are on MATH-500. JSD on Qwen3-8B has a 13.1 pp degradation and Forward KL on 
Olmo-3-7B-Instruct also has a 6.6 pp degradation.
One explanation is that reverse KL weights each entry by the student probability and tends to concentrate the student on the modes of the teacher, which suits a teacher whose context already led to a correct answer.

\section{Related Work}

\paragraph{Exploration in RLVR.}
Sampling a group of independent responses for each prompt is a fundamental approach to estimating advantages in RLVR~\citep{shao2024deepseekmath,yu2025dapo,zheng2025gspo}. Prior work finds that the efficiency of RLVR depends on which solutions are explored under a limited verification budget~\citep{zhang2026barl,setlur2026e3,wu2026reasoningcache}. Larger groups allow for more extensive exploration that can substantially improve the estimate. 
However, sampling in this way suffers from a fundamental limitation: every response is generated from the same problem prompt, so a failed attempt cannot inform the next sample.
Our work therefore aims to condition each rollout on earlier attempts at the same problem, which brings new information in every rollout trace.

\paragraph{Feedback guided refinement.}
Prior work explores prompting LLMs to revise their own outputs using information from earlier attempts. \citet{shinn2023reflexion} stores verbal reflection across interactions. \citet{madaan2023selfrefine} introduces iterative self-refinement. \citet{song2026icrl,yu2026icpo} leverage in-context reinforcement learning and policy optimization.
Some methods consider training models to use feedback more effectively rather than relying on prompting alone. They improve through textual critiques, explicit reflection or retrospective credit assignment~\citep{du2025ptr,song2026rltf,shi2026erl,li2026mulferl,ekbote2025murphy,jin2026reveal,jain2025mucode,yuan2025dpsdp}.
Such training may occur over multiple turns, such as \citet{kumar2024score}, which trains models with reinforcement learning on two turns of self-correction. \citet{mu2026drift} decouple multi-turn rollouts from a weighted fine-tuning by importance.
In contrast to these works, \ours{} generates one rollout per problem per step and updates the parameters by self-distillation on single trajectories, without reward objectives on groups of answers.

\paragraph{Self-distillation with privileged context.}
The memory-conditioned policy in \ours{} can be viewed as a teacher with privileged context that is updated as the policy explores.
Prior work primarily considers distilling a richer context into a model that must eventually answer without it~\citep{snell2022context,agarwal2024gkd}, where the context holds environment feedback~\citep{hubotter2026sdpo}, privileged information on student-generated trajectories~\citep{zhao2026opsd}, self-revision~\citep{he2026sdzero}, procedural memories~\citep{liu2026pmd} or skills~\citep{wang2026skillsd,jiang2026skald}.
Others combine distillation with supervised fine-tuning, preference optimization or reinforcement learning~\citep{yang2025supercorrect,wu2026seed}. 

\section{Conclusion}

Independent rollouts make exploration less efficient in LLM reinforcement learning. We propose \ours{}, which keeps a dynamic memory for each rollout trace and updates the policy by self-distillation, to improve learning efficiency. Our experiments found that \ours{} reached the final pass rate of Random Sampling with about four times fewer verifier calls and ended above all RL baselines at lower compute without policy update. The self-distillation further gave the best overall test accuracy. We conducted several ablation studies on key designs. The results highlight that \ours{} can be a new framework with better exploration efficiency.

\section*{Limitations}

On LiveCodeBench and MBPP+ the feedback includes execution results such as errors and failed tests, while on MATH-500 and SciKnowEval it is only a correctness verdict, so the effect of richer feedback is tested on code tasks only. Although we conducted ablation studies on key components, how memory affects exploration still needs further research. 

\section*{Ethical Considerations}
\label{sec:ethics}

We use public datasets and open-weight models without collecting new personal data or involving human participants. We rely on the original dataset curation without additional screening for personal information or offensive content.

\ours{} may inherit model biases and propagate errors from verifier feedback, including judgments from the LLM judge. Benchmark gains do not establish deployment safety: scientific answers require independent verification, and generated code requires security review. Computational costs are reported in Appendices~\ref{app:explore-setup} and~\ref{app:general-setup}.

\bibliography{references}

\appendix

\begin{table}[b!]
\centering
\small
\setlength{\tabcolsep}{4pt}
\begin{tabular}{l r r r r}
\toprule
& \multicolumn{2}{c}{Qwen3-8B} & \multicolumn{2}{c}{Olmo-3-7B-Instruct} \\
\cmidrule(lr){2-3}\cmidrule(lr){4-5}
Benchmark & Wrong & Repeat (\%) & Wrong & Repeat (\%) \\
\midrule
MATH-500 & 1,912 & 45.7 & 1,309 & 47.4 \\
LCB & 2,480 & 61.8 & 2,452 & 40.5 \\
MBPP+ & 1,278 & 77.3 & 1,675 & 78.0 \\
\bottomrule
\end{tabular}
\caption{Repeated wrong answers among the 24 independent Random Sampling responses per problem. Wrong is the total number of wrong responses over all problems. Repeat is the fraction of wrong responses whose error signature equals that of an earlier wrong response to the same problem. Correctness on LCB and MBPP+ is judged on public tests.}
\label{tab:repeat-random}
\end{table}

\begin{table*}[t]
\centering
\small
\setlength{\tabcolsep}{3.5pt}
\resizebox{\textwidth}{!}{%
\begin{tabular}{l l l r r r r r}
\toprule
Model & Benchmark & Comparison & Problems & Repeat (\%) & $\Delta$ Repeat (pp) & Fix (\%) & $\Delta$ Fix (pp) \\
\midrule
\multirow{6}{*}{Qwen3-8B}
& \multirow{2}{*}{MATH-500} & w/o feedback & 143 & 52.5 / 65.0 & $-12.5$ ($-17.9$, $-7.4$) & 29.5 / 19.8 & $+9.7$ ($+4.2$, $+15.1$) \\
& & w/o last response & 148 & 51.6 / 12.4 & $+39.2$ ($+34.1$, $+44.4$) & 30.9 / 53.0 & $-22.2$ ($-28.0$, $-16.7$) \\
& \multirow{2}{*}{LCB} & w/o feedback & 115 & 38.3 / 67.6 & $-29.4$ ($-35.8$, $-23.3$) & 16.6 / 3.5 & $+13.0$ ($+8.2$, $+18.1$) \\
& & w/o last response & 117 & 38.2 / 23.8 & $+14.4$ ($+9.6$, $+18.9$) & 17.3 / 16.0 & $+1.3$ ($-1.7$, $+4.4$) \\
& \multirow{2}{*}{MBPP+} & w/o feedback & 114 & 36.5 / 79.2 & $-42.7$ ($-49.0$, $-36.0$) & 54.6 / 16.7 & $+37.9$ ($+30.9$, $+44.9$) \\
& & w/o last response & 126 & 36.0 / 27.3 & $+8.7$ ($+2.8$, $+15.1$) & 55.4 / 60.5 & $-5.1$ ($-11.9$, $+1.2$) \\
\midrule
\multirow{6}{*}{Olmo-3-7B-Instruct}
& \multirow{2}{*}{MATH-500} & w/o feedback & 72 & 41.1 / 52.8 & $-11.7$ ($-20.0$, $-3.3$) & 40.1 / 31.1 & $+9.0$ ($+1.0$, $+16.9$) \\
& & w/o last response & 77 & 39.0 / 13.9 & $+25.1$ ($+18.0$, $+32.7$) & 44.0 / 53.8 & $-9.8$ ($-17.3$, $-3.2$) \\
& \multirow{2}{*}{LCB} & w/o feedback & 123 & 24.1 / 43.4 & $-19.3$ ($-24.0$, $-14.8$) & 23.6 / 15.5 & $+8.1$ ($+3.9$, $+12.4$) \\
& & w/o last response & 122 & 24.3 / 9.3 & $+15.0$ ($+11.3$, $+18.6$) & 23.0 / 18.4 & $+4.6$ ($+0.7$, $+8.8$) \\
& \multirow{2}{*}{MBPP+} & w/o feedback & 90 & 22.8 / 68.5 & $-45.7$ ($-52.8$, $-38.6$) & 64.7 / 25.8 & $+38.9$ ($+30.7$, $+47.0$) \\
& & w/o last response & 112 & 20.9 / 25.0 & $-4.1$ ($-11.0$, $+3.0$) & 68.5 / 59.1 & $+9.3$ ($+1.0$, $+17.4$) \\
\bottomrule
\end{tabular}}
\caption{Repeat and fix rates of \ours{} compared with two memory ablations over the first 24 exploration steps. Repeat and Fix entries are listed as \ours{} / comparison, averaged per problem over transitions and then over the problems shared by both conditions (Problems). $\Delta$ is \ours{} minus the comparison in percentage points, with the 95\% bootstrap confidence interval over problems in parentheses.}
\label{tab:repeat-paired}
\end{table*}

\begin{table*}[t]
\centering
\small
\setlength{\tabcolsep}{4.5pt}
\begin{tabular}{l l r r r r r r}
\toprule
& & \multicolumn{3}{c}{\ours{}} & \multicolumn{3}{c}{\ours{} w/o feedback} \\
\cmidrule(lr){3-5}\cmidrule(lr){6-8}
Model & Benchmark & Repeat (\%) & Base (\%) & Ratio & Repeat (\%) & Base (\%) & Ratio \\
\midrule
\multirow{3}{*}{Qwen3-8B}
& MATH-500 & 60.9 & 16.6 & 3.67 & 73.4 & 17.3 & 4.24 \\
& LCB & 41.8 & 23.3 & 1.79 & 66.8 & 30.2 & 2.21 \\
& MBPP+ & 55.4 & 40.9 & 1.35 & 85.0 & 38.1 & 2.23 \\
\midrule
\multirow{3}{*}{Olmo-3-7B-Instruct}
& MATH-500 & 56.0 & 21.0 & 2.67 & 65.0 & 23.8 & 2.73 \\
& LCB & 27.4 & 9.7 & 2.84 & 47.1 & 10.5 & 4.47 \\
& MBPP+ & 36.4 & 38.2 & 0.95 & 78.9 & 44.8 & 1.76 \\
\bottomrule
\end{tabular}
\caption{Repeat rates relative to independent sampling, computed over all transitions of each condition together. Base is the fraction of the 24 Random Sampling responses to the same problem that are wrong with the same error signature as the previous answer, averaged over transitions. Ratio is the number of repeats divided by the sum of base frequencies.}
\label{tab:repeat-base}
\end{table*}

\section{Repeated wrong answers under independent sampling}
\label{app:repeat}

This section specifies how repeated wrong answers are measured and reports the per-benchmark measurements behind the repeat statistics in the Introduction. We analyze MATH-500, LCB and MBPP+ for Qwen3-8B and Olmo-3-7B-Instruct. Unless stated otherwise, all runs in this section are single runs with the data, budget and decoding settings of Table~\ref{tab:explore}.

\paragraph{Error signatures.}
We introduce error signatures to identify repetitive answers. On MATH-500, the error signature of a wrong answer is its normalized final answer. On LCB and MBPP+, the signature is the full public-test feedback text, that is, the failing input, output or error message shown to the model. Throughout this section, correctness on LCB and MBPP+ is judged on the public tests only.

\paragraph{Repeats under independent sampling.}
For each problem, we take the 24 Random Sampling responses in the order in which they were drawn. A wrong response counts as a repeat if its signature equals the signature of an earlier wrong response to the same problem. Wrong responses without an extracted answer count as non-repeats. The repeat rate is the number of repeats divided by the number of wrong responses, both summed over problems. Table~\ref{tab:repeat-random} reports these rates.

\paragraph{Memory conditions and transitions.}
To measure repeats when the previous attempt is in the context, we compare three exploration conditions that differ only in the memory. \ours{} shows the previous response and its feedback. \ours{} w/o feedback shows the previous response followed by the blind correction instruction, which contains no correctness signal or test output. \ours{} w/o last response shows the problem and the feedback in a single user turn, without the previous response. These are the single runs of the ablations in Figure~\ref{fig:ablation}, restricted to the first 24 steps. A transition is a step $t$ that is not a restart step and whose previous step $t-1$ produced a wrong answer with an error signature. With $M=5$, the restart steps are $t=1, 6, 11, 16, 21$. These steps are excluded because the policy revises a fresh draft from the plain prompt instead of the step-$(t-1)$ answer.

\paragraph{Repeat and fix rates.}
Over transitions, the repeat rate is the fraction in which step $t$ is wrong with the same signature as step $t-1$. The fix rate is the fraction in which step $t$ is correct. For the paired comparison in Table~\ref{tab:repeat-paired}, we first average each problem's repeat and fix rates over its transitions. We then average over the problems that have at least one transition under both \ours{} and the comparison condition.

Table~\ref{tab:repeat-paired} reports the paired comparisons. Relative to \ours{} w/o feedback, \ours{} has a lower repeat rate and a higher fix rate in all six pairs, and its repeat rate averaged over the six pairs is lower by about 27 pp. Relative to \ours{} w/o last response, \ours{} has a higher repeat rate in five of the six pairs and a lower fix rate on MATH-500 for both models.

\paragraph{Base frequency under independent sampling.}
To relate the repeat rates of \ours{} to independent sampling, we use the 24 Random Sampling responses to the same problem. For each transition, the base frequency is the fraction of those 24 responses that are wrong with the same signature as the step-$(t-1)$ answer. Table~\ref{tab:repeat-base} pools all transitions of a condition, without the problem matching used in Table~\ref{tab:repeat-paired}. It reports the average repeat rate, the base frequency averaged over transitions, and the ratio of the number of repeats to the sum of base frequencies.

Table~\ref{tab:repeat-base} compares these repeat rates with the base frequency under independent sampling. The repeat rates differ from those in Table~\ref{tab:repeat-paired} because they weight each transition equally and are not restricted to shared problems. The ratio for \ours{} is lower than that for \ours{} w/o feedback in all six pairs. Olmo-3-7B-Instruct MBPP+ is the only cell where the ratio for \ours{} is below 1, so there \ours{} repeats a wrong answer less often than independent sampling.

\section{Method details}
\label{app:method}

This section specifies the memory context, the restart step, the prompts and the policy update of \ours{} in Algorithm~\ref{alg:ours}. Unless stated otherwise, the exploration runs (Table~\ref{tab:explore}) and the rollouts on the training sets that supply the teacher contexts for the policy update (Table~\ref{tab:general}) use the same construction. Benchmarks, decoding parameters and budgets are given in Appendix~\ref{app:explore-setup} and Appendix~\ref{app:general-setup}.

\subsection{Memory context construction}

\paragraph{Dialogue format.}
At a carry step $t$, meaning a step that is not a restart, the context $c_t$ is a chat dialogue with three parts. The first part is the opening prompt of the problem. It is the user turn given below, preceded by a system turn for Olmo-3-7B-Instruct on LCB. The second part is an assistant turn that holds $\hat y_{t-1}=\tail_L(y_{t-1})$. The third part is a user turn made of the feedback section of step $t-1$ and an improvement instruction, separated by a blank line. The assistant turn holds only the previous response, so earlier attempts in the same restart segment are not shown. For Qwen3-8B, we apply the chat template with thinking mode disabled. Olmo-3-7B-Instruct uses its own chat template, which ignores this flag.

\paragraph{Rendering of verdict and feedback.}
The feedback section depends on the verifier of each benchmark. On LCB and MBPP+, it starts with a sentence stating that the previous answer was executed against the public tests. A verdict sentence follows, and then the native feedback of the executor. When the executor returns no details, a fixed fallback sentence is used in their place. On MATH-500 and SciKnowEval, the section has only a sentence stating that the answer was checked against the reference, followed by the verdict. On MATH-500, the extracted final answer is not quoted back. On SciKnowEval, neither the judge's reasoning nor the reference answer is shown.

On LCB, the native feedback comes from the SDPO code executor and covers failing tests only. If any test raises a runtime error, only the first such test is shown. Otherwise, the first timeout is shown, and failing that, the first test with an incorrect format. If none of these occur, the wrong answers are sorted by the combined length of the input and the actual output, shortest first, and at most two are shown. A wrong answer is rendered as the test index, the input, the actual output and the expected output. A runtime error is rendered as the error message and the last executed input. Each input line is capped at 250 characters and each input at 8 lines. The expected and actual outputs are each capped at 250 characters, and the whole feedback is capped at 2,000 characters. Tracebacks keep at most 3 frames, all of them inside the submitted solution. The program is taken from the longest fenced code block. On the LCB training set used for the policy update, tests run in order and stop at the first failure, so the feedback shows that one failure.

On MBPP+, each of the three public asserts is run and reported separately. The feedback starts with ``Public test results: $p$/$n$ asserts passed.'' It then lists each failed assert together with its exception, capped at 300 characters. When all asserts pass, the feedback is ``All public asserts passed.'' Separate messages cover four cases: a program that does not compile, an execution timeout after 30 seconds, an exception raised before any test runs, and an answer from which no program can be extracted. The whole feedback is capped at 2,000 characters. MATH-500 answers are graded with math-verify~\citep{kydlicek2025mathverify} against the reference answer. SciKnowEval answers are graded by an LLM judge, which is the run's own base model without LoRA.

\paragraph{Tail truncation.}
The transform $\tail_L$ uses $L=2{,}000$ tokens. If a response has more than 2,000 tokens, the assistant turn holds the decoded text of its last 2,000 token IDs, with special tokens skipped. Otherwise the response is kept verbatim. The same transform is applied to the restart draft $\tilde y$.

\subsection{Restart with blind revision}

A restart occurs at step 1 and every $M=5$ steps after it. Within the budget of $K=24$ calls, steps 1, 6, 11, 16 and 21 are restarts. At a restart step, the policy first drafts $\tilde y\sim\pi_\theta(\cdot\mid x)$ from the opening prompt. The context then becomes the opening prompt, an assistant turn holding $\tail_L(\tilde y)$, and a user turn holding the blind correction instruction for the task type. The policy generates $y_t$ under this context. Only $y_t$ is sent to the verifier and counted toward the budget, and the draft is not graded. The blind instruction contains no verdict and no test output. The next carry step uses $y_t$ and its feedback as memory.

\subsection{Exact prompts}

Each prompt below is the full text of one turn.

\paragraph{Opening prompts for Qwen3-8B.}
The user turn is the benchmark instruction, a blank line and the problem text. There is no system turn. The LCB instruction is:
\begin{quote}\small\ttfamily
You are a coding expert. You will be given a coding problem, and you need to write a correct Python program that matches the specification and passes all tests. The time limit is 1 second. You may start by outlining your thought process. In the end, please provide the complete code in a code block enclosed with \textasciigrave\textasciigrave\textasciigrave{} \textasciigrave\textasciigrave\textasciigrave.
\end{quote}
The MATH instruction is:
\begin{quote}\small\ttfamily
You are a mathematics expert. You will be given a competition mathematics problem. Solve it and give the final answer. You may start by outlining your thought process. In the end, please put the final answer inside \textbackslash boxed\{\}.
\end{quote}
The MBPP+ instruction is:
\begin{quote}\small\ttfamily
You are a coding expert. You will be given a Python programming task and the tests your code must satisfy. Write a correct Python program that matches the specification and passes all tests. Keep the function names and signatures the tests require. You may start by outlining your thought process. In the end, please provide the complete code in a code block enclosed with \textasciigrave\textasciigrave\textasciigrave{} \textasciigrave\textasciigrave\textasciigrave.
\end{quote}
The MBPP+ problem text is the task description, followed by ``Your code should pass these tests:'' and the three public asserts. For a task whose asserts need setup code, that code follows the asserts. The SciKnowEval instruction is:
\begin{quote}\small\ttfamily
You are a scientific reasoning expert. You will be given a science task. Reason about it step by step, then give the final answer. In the end, write "Final answer:" followed by your complete final answer, in the format the task asks for.
\end{quote}
The SciKnowEval problem text is the dataset's default task prompt, a blank line and the question.

\paragraph{Opening prompts for Olmo-3-7B-Instruct.}
Olmo-3-7B-Instruct uses the OLMES templates. On MATH, the user turn is the problem followed by a blank line and:
\begin{quote}\small\ttfamily
Present the answer in LaTex format: \textbackslash boxed\{Your answer\}
\end{quote}
On LCB, the system turn is:
\begin{quote}\small\ttfamily
You are an expert Python programmer. You will be given a question (problem specification) and will generate a correct Python program that matches the specification and passes all tests.
\end{quote}
The LCB user turn is:
\begin{quote}\small\ttfamily
\#\#\# Question:\par
\{question\_content\}\par
\medskip
\#\#\# Format:\par
Provide CONCISE reasoning on how to arrive at the answer in the <think> </think> tag.\par
\{format\}\par
\medskip
\#\#\# Answer: (use the provided format with backticks)
\end{quote}
Here, \texttt{\{format\}} takes one of two forms. With starter code, it is ``You will use the following starter code to write the solution to the problem and enclose your code within delimiters.'' followed by the starter code in a Python block. Without starter code, it is ``Read the inputs from stdin solve the problem and write the answer to stdout (do not directly test on the sample inputs). Enclose your code within delimiters as follows. Ensure that when the python program runs, it reads the inputs, runs the algorithm and writes output to STDOUT.'' followed by a Python block containing \texttt{\# YOUR CODE HERE}. The MBPP+ user turn is:
\begin{quote}\small\ttfamily
\{description\}\par
\{function header\}\par
\medskip
Your code should pass these tests:\par
\medskip
\{three asserts\}\par
\medskip
Provide CONCISE reasoning on how to arrive at the answer, and make sure to finish the response with the following:\par
\medskip
Here is the completed function:\par
\medskip
\textasciigrave\textasciigrave\textasciigrave python\par
(CODE)\par
\textasciigrave\textasciigrave\textasciigrave\par
where (CODE) is the code for the complete function.
\end{quote}
The SciKnowEval user turn is:
\begin{quote}\small\ttfamily
Answer the following science task. Provide CONCISE reasoning for the answer, and make sure to finish the response with "Final answer:" followed by your complete answer in the format the task asks for.\par
\medskip
Task: \{problem\}\par
\medskip
Answer the above task and REMEMBER to finish your response with "Final answer:" followed by your complete answer.
\end{quote}
Here, \texttt{\{problem\}} is the same problem text as for Qwen3-8B.

\paragraph{Carry-step user turn.}
On LCB and MBPP+, the final user turn of a carry step is:
\begin{quote}\small\ttfamily
Your previous answer was executed against the public tests.\par
\medskip
\{verdict\}\par
\medskip
\{native\_feedback\}\par
\medskip
Use this execution feedback to improve your answer. Re-check the original problem yourself and produce a corrected complete answer.
\end{quote}
Here, \texttt{\{verdict\}} is ``Your previous answer passed every public test.'' or ``Your previous answer failed the public tests.'' When the native feedback is empty, it is replaced by ``No failing test details were produced.'' if the answer passed and by ``No failing test details are available.'' otherwise. On MATH-500 and SciKnowEval, the final user turn is:
\begin{quote}\small\ttfamily
Your previous answer was checked against the reference.\par
\medskip
\{verdict\}\par
\medskip
Use this outcome to improve your answer. Re-check the original problem yourself and produce a corrected complete answer.
\end{quote}
Here, \texttt{\{verdict\}} is ``Your previous answer was judged correct.'' or ``Your previous answer was judged incorrect.''

\paragraph{Blind correction instructions.}
At a restart step, the user turn after the draft holds the blind instruction for the task type. For code tasks (LCB and MBPP+), it is:
\begin{quote}\small\ttfamily
Review your previous answer carefully for reasoning, algorithmic, complexity, and implementation mistakes. Then produce a corrected complete answer to the original problem. Do not assume that any external correctness signal or test feedback is available.
\end{quote}
For MATH, it is:
\begin{quote}\small\ttfamily
Review your previous answer carefully for errors in reasoning, algebra, arithmetic, and the handling of special or boundary cases. Then produce a corrected complete answer to the original problem. Do not assume that any external correctness signal is available.
\end{quote}
For SciKnowEval, it is:
\begin{quote}\small\ttfamily
Review your previous answer carefully for errors in scientific reasoning, the facts and relationships you relied on, unit and magnitude handling, and whether you addressed what the question actually asks. Then produce a corrected complete answer to the original problem. Do not assume that any external correctness signal is available.
\end{quote}

\paragraph{SciKnowEval judge.}
The judge reads the text after the last \texttt{</think>} of the response if that text is non-empty, and the whole response otherwise. A response longer than 12,000 characters is shown as its first 3,000 and last 9,000 characters, with an omission marker between them. The judge user message is:
\begin{quote}\small\ttfamily
You are an expert grader for science tasks. Decide whether the response to the task below is correct, using the reference answer as the ground truth.\par
\medskip
\#\#\# Task\par
\{question\}\par
\medskip
\#\#\# Reference answer\par
\{reference\}\par
\medskip
\#\#\# Response\par
\{response\}\par
\medskip
\#\#\# Grading instructions\par
\{criterion\}\par
Judge the final answer the response gives; its style and length do not matter. Be strict but fair. A response that gives no final answer, hedges between several answers, or is cut off before answering is incorrect.\par
\medskip
Explain your judgment in at most three sentences, then end with \textbackslash boxed\{yes\} if the response is correct or \textbackslash boxed\{no\} if it is not.
\end{quote}
Here, \texttt{\{criterion\}} is one of six criterion texts, selected by the SciKnowEval subtask. The verdict is read from the last \texttt{\textbackslash boxed\{yes\}} or \texttt{\textbackslash boxed\{no\}} in the judge output, and any other output counts as incorrect. The judge decodes greedily (temperature 0) with at most 512 new tokens.

\subsection{Self-distillation implementation}

\paragraph{Teacher contexts.}
Before the policy update, \ours{} runs $K=20$ steps of the rollout above on each training set (LCB, MATH and SciKnowEval). For each problem, $\mathcal{C}^{+}(x)$ holds the contexts whose final response was correct and not truncated. A response that exactly repeats an earlier correct response to the same problem is dropped. A context is the full chat dialogue that produced the response. At a carry step, this dialogue includes the earlier attempt and its feedback. At a restart step, it includes the draft and the blind instruction. Only problems with a non-empty $\mathcal{C}^{+}(x)$ enter the update. At each update step, one context is drawn uniformly from $\mathcal{C}^{+}(x)$ for each problem in the batch. 

\paragraph{Batches and student samples.}
The update runs $S=20$ steps with 128 problems per step and one optimizer step per training step. Batches are cut from a fixed stream that concatenates per-epoch permutations of the eligible problems, where the permutation of epoch $e$ is seeded by the schedule seed plus $e$. For each problem, one student response $y\sim\pi_\theta(\cdot\mid x)$ is sampled with vLLM~\citep{kwon2023vllm} from the plain opening prompt. 

\paragraph{Teacher distribution.}
The teacher is the current policy with its current LoRA adapter. The teacher reads the drawn context followed by the same student response. Its logits are computed without gradients, which implements $\sg$ in the definition of $q_i$. Student and teacher logits are taken at every response position and divided by the model's sampling temperature. The divergence is computed over the full vocabulary.

\paragraph{Importance weights.}
The student response is sampled with vLLM, whose token probabilities $\mu$ can differ slightly from those of the training engine. Before the update step, the training engine recomputes the log probability of every sampled token under the current parameters, and $w^{x}_{i}=\min\big(\pi_\theta(y^{x}_{i}\mid x,y^{x}_{<i})/\mu(y^{x}_{i}\mid x,y^{x}_{<i}),\,2\big)$ in Equation~\ref{eq:loss}, where $\mu$ is the vLLM probability of the sampled token. The weights are constants in the loss.

\paragraph{Clipped $f$-divergence.}
With the default $f(u)=u\log u$, each summand of Equation~\ref{eq:fdiv} is $p_{i,v}\log(p_{i,v}/q_{i,v})$, so the token loss is
\begin{equation}
D^{\tau}(p_i\,\|\,q_i)=\sum_{v\in\mathcal{V}}\min\!\left(p_{i,v}\log\frac{p_{i,v}}{q_{i,v}},\,\tau\right).
\label{eq:app-clipped-kl}
\end{equation}
We use $\tau=10^{-3}$ on MATH. On LCB and SciKnowEval, no cap is applied, and the loss is the unclipped reverse KL $\mathrm{KL}(p_i\,\|\,q_i)$. The forward-KL and JSD variants in Section~\ref{sec:loss} replace only the summand and keep the same cap on MATH.

\paragraph{Parameterization and optimization.}
We train a LoRA adapter with rank 64, alpha 64 and dropout 0.0 on all linear layers, with the backbone frozen. We use AdamW~\citep{loshchilov2019adamw} with betas $(0.9, 0.999)$, weight decay 0.01, a maximum gradient norm of 1.0 and a constant learning rate of $10^{-5}$ with no warmup. Training runs in bfloat16 with FSDP~\citep{zhao2023fsdp}, gradient checkpointing, dynamic batch sizes and padding removal. Each run uses one H200 GPU and trains a separate adapter for each training set.

\section{Exploration setup details}
\label{app:explore-setup}

This section specifies the datasets, verifiers, cost measurement, and baseline configurations underlying the exploration results in Table~\ref{tab:explore} and Figure~\ref{fig:explore}. Unless stated otherwise, every method is run on both Qwen3-8B with thinking disabled and Olmo-3-7B-Instruct, receives $K=24$ verifier calls per problem, and is limited to 5 GPU hours per benchmark on one H200 GPU. The prompts that build the memory dialogue of \ours{} are given in Appendix~\ref{app:method}.

\subsection{Datasets}

The exploration pool contains 1,409 problems: 500 from MATH-500, 131 from LCB, 378 from MBPP+, and 400 from SciKnowEval. Each benchmark is trained and evaluated separately. The problem order within each benchmark file is shuffled with a fixed seed.

\paragraph{MATH-500.} We use all 500 problems without filtering. Answers are graded with math-verify against the last \texttt{\textbackslash boxed\{\}} expression in the response.

\paragraph{LCB.} We use the 131 problems from release v6 of LiveCodeBench with contest dates from 2025-02-01 to 2025-04-30, which is the split used by SDPO. The test cases of each problem are the private test cases released with LiveCodeBench. The worked examples in the problem statement belong to neither the public nor the private set defined below.

\paragraph{MBPP+.} We use all 378 tasks of EvalPlus~\citep{liu2023evalplus} MbppPlus v0.2.0. The prompt contains the task description followed by the three assert statements that the code must satisfy.

\paragraph{SciKnowEval.} We start from the v2 test set and keep only the \texttt{open-ended-qa} and \texttt{relation\_extraction} types at levels L2 and L5, which excludes all multiple-choice and true/false problems. We then drop entries whose (instruction, question) pair occurs more than once. For each of the four domains (Biology, Chemistry, Material, and Physics), we draw 75 problems from L2 and 25 from L5 with a fixed seed, giving 100 problems per domain and 400 in total.

\subsection{Public and private tests}

On the two coding benchmarks, the verifier reward and the feedback shown to the policy use only public tests. A response counts as correct for the pass rate only if it passes both the public and the private tests (Equation~\ref{eq:pass}). For LCB, we split the test cases of each problem with a fixed random seed: half of the cases form the public set, and the remaining cases form the private set. Public tests run in the SDPO executor. Private tests are graded offline with the official LiveCodeBench runner. For MBPP+, the public set is the three asserts shown in the prompt, and each assert is executed separately. The private set is the MBPP+ \texttt{plus\_input} cases.

\subsection{Verifiers and feedback}

\paragraph{Coding feedback.} On LCB, the feedback lists failing public tests only. If a test raises a runtime error, the feedback shows the first such test with its error message and input. Otherwise, it shows the first timeout, then an incorrect-format message, and otherwise at most two wrong answer cases, chosen as the shortest by combined input and output length, each with its input, actual output, and expected output. Truncated responses receive a fixed message asking for a shorter response. Long inputs and outputs are truncated, tracebacks keep at most three frames inside the solution, and the whole feedback is capped at 2,000 characters. On MBPP+, the feedback reports the number of passed asserts and, for each failed assert, the assert and its exception. It also reports compilation errors, timeouts, and responses from which no program can be extracted.

\paragraph{Verdict-only feedback.} On MATH-500 and SciKnowEval, the feedback states only whether the previous answer was judged correct. On MATH-500, neither the reference answer nor the extracted final answer is shown.

\paragraph{LLM judge.} SciKnowEval answers are graded by the base model of the run without LoRA. The judge decodes greedily (temperature 0, top-$p$ 1) with at most 512 new tokens. It sees only the text after the last \texttt{</think>} tag. A response longer than 12,000 characters is shown as its first 3,000 and last 9,000 characters, with an omission marker in between. The verdict is the last \texttt{\textbackslash boxed\{yes\}} or \texttt{\textbackslash boxed\{no\}} in the judge output, and any other output counts as incorrect. The judge user message is:
\begin{quote}\small\ttfamily
You are an expert grader for science tasks. Decide whether the response to the task below is correct, using the reference answer as the ground truth.\par
\medskip
\#\#\# Task\par
\{question\}\par
\medskip
\#\#\# Reference answer\par
\{reference\}\par
\medskip
\#\#\# Response\par
\{response\}\par
\medskip
\#\#\# Grading instructions\par
\{criterion\}\par
Judge the final answer the response gives; its style and length do not matter. Be strict but fair. A response that gives no final answer, hedges between several answers, or is cut off before answering is incorrect.\par
\medskip
Explain your judgment in at most three sentences, then end with \textbackslash boxed\{yes\} if the response is correct or \textbackslash boxed\{no\} if it is not.
\end{quote}
The \texttt{\{criterion\}} field is one of six texts selected by the SciKnowEval subtask: relation set, composition, summary, key content, derivation, and final result. The final-result criterion is:
\begin{quote}\small\ttfamily
The response is correct if its final result is equivalent to the reference answer: the same formula, value, expression, structure or material, allowing equivalent notation, units and rounding. The reasoning does not need to match the reference. If the reference answers several parts, every part must match.
\end{quote}
The verifier for each benchmark, including the judge, is the same for every method.

\begin{table*}[t]
\centering
\small
\setlength{\tabcolsep}{3.5pt}
\resizebox{\textwidth}{!}{%
\begin{tabular}{l c c c c c c}
\toprule
Setting & Random Sampling & GRPO & SDPO & SCoRe & SEED & \ours{} \\
\midrule
Verifier calls / problem & 24 & 24 & 24 & 24 & 24 & 24 \\
GPU-hour cap / benchmark & 5 & 5 & 5 & 5 & 5 & 5 \\
Parameter updates & \xmark & \cmark & \cmark & \cmark & \cmark & \xmark \\
Questions / update & -- & 32 & 32 & 32 & 32 & -- \\
Responses / question & 1 per step & 8 & 8 & $4\times2$ turns & 8 & \makecell{1 verified per step\\(+1 draft at restart)} \\
Updates & -- & 47/13/36/38 & 47/13/36/38 & 47/13/36/38 & \makecell{8/2/6/6 (collect)\\+ 39/11/30/32 (RL)} & -- \\
LoRA rank / alpha & -- & 64 / 64 & 64 / 64 & 64 / 64 & 64 / 64 & -- \\
Learning rate & -- & $10^{-6}$ & $10^{-6}$ & $10^{-6}$ & $10^{-6}$ & -- \\
Clip ratio (low / high) & -- & 0.2 / 0.28 & -- & 0.2 & 0.2 & -- \\
KL coefficient to reference & -- & none & -- & 1.0 & 0.01 & -- \\
Response limit (tokens) & 8,192 & 8,192 & 8,192 & 8,192 per turn & 8,192 & 8,192 \\
Other & -- & -- & \makecell{top-20 + tail\\EMA rate 0.01} & \makecell{$\beta_1=0.01$, $\beta_2=0.25$\\bonus 10.0} & \makecell{SFT 3 epochs\\gate 5.0, OPD 0.01} & \makecell{$M=5$\\$L=2{,}000$} \\
\bottomrule
\end{tabular}}
\caption{Exploration hyperparameters for all methods. The response limit is 16,384 tokens for Olmo-3-7B-Instruct on LCB, and SCoRe limits the two turns to 16,896 tokens in total. All methods with parameter updates use AdamW with weight decay 0.01, a maximum gradient norm of 1.0, and a constant learning rate. Dashes mark settings that do not apply or are not set separately for that method.}
\label{tab:explore-hparams}
\end{table*}

\subsection{Metrics and compute measurement}

\paragraph{Aligned steps.} Step $k$ of a problem refers to its first $k$ graded answers. Random Sampling and \ours{} produce one answer per problem per step. For the batch methods (GRPO, SDPO, SCoRe, and SEED), we align on the $k$-th answer of each problem rather than on the $k$-th training batch. The cumulative pass rate at step $k$ follows Equation~\ref{eq:pass}. The Pass column of Table~\ref{tab:explore} is the mean of the four per-benchmark pass rates at the last aligned step.

\paragraph{Budget cutoff.} Each run stops at $\min(24, k_{5\mathrm{h}})$ steps, where $k_{5\mathrm{h}}$ is the last step whose cumulative GPU time does not exceed 5 hours. A run cut by this cap is evaluated at the steps it completed.

\paragraph{GPU hours.} All jobs run on a single H200 GPU (141 GB). For \ours{}, we take the measured rollout wall time, split it over steps in proportion to the per-problem sampling time of each step, and add the start-up overhead. For the methods based on verl~\citep{sheng2025hybridflow}, the time of each step is the initial overhead plus each answer's share of the measured update time. For Random Sampling, the measured time of each generation chunk is allocated to steps by token count. The GPU hours in Table~\ref{tab:explore} are the cumulative values at the last aligned step, summed over the four benchmarks.

\paragraph{FLOPs.} We estimate compute as
\begin{equation}
\mathrm{FLOPs}=2N\cdot T_{\mathrm{fwd}}+6N\cdot T_{\mathrm{train}},
\label{eq:flops}
\end{equation}
where $N=8.19\times10^{9}$ for Qwen3-8B and $7.30\times10^{9}$ for Olmo-3-7B-Instruct. Here, $T_{\mathrm{fwd}}$ is the number of tokens processed in forward passes, and $T_{\mathrm{train}}$ is the number of tokens that receive gradients. For \ours{} and Random Sampling, $T_{\mathrm{fwd}}$ counts the prompt and completion tokens of every generation call, including the restart drafts of \ours{} and every judge call, and $T_{\mathrm{train}}=0$. For the methods based on verl, $T_{\mathrm{fwd}}$ counts the generated sequences, the old policy and reference log-probability passes, and the judge calls, together with the SDPO teacher forward pass, the second SCoRe turn, and the SEED analyzer and teacher passes. $T_{\mathrm{train}}$ is the number of tokens in the policy updates. For SEED, $T_{\mathrm{fwd}}$ also includes the collection rollouts and analyses, and $T_{\mathrm{train}}$ includes three epochs of SFT tokens. Table~\ref{tab:explore} reports the result in PFLOPs ($10^{15}$ FLOPs).

\begin{table*}[t]
\centering
\small
\setlength{\tabcolsep}{3.5pt}
\resizebox{\textwidth}{!}{%
\begin{tabular}{llrrrrrrrrrr}
\toprule
 & & \multicolumn{2}{c}{MATH-500} & \multicolumn{2}{c}{LCB} & \multicolumn{2}{c}{MBPP+} & \multicolumn{2}{c}{SciKnowEval} & \multicolumn{2}{c}{Overall} \\
\cmidrule(lr){3-4}\cmidrule(lr){5-6}\cmidrule(lr){7-8}\cmidrule(lr){9-10}\cmidrule(lr){11-12}
Model & Method & Acc. & Pass & Acc. & Pass & Acc. & Pass & Acc. & Pass & Acc. & Pass \\
\midrule
\multirow{6}{*}{Qwen3-8B}
 & Random Sampling & 84.14 & 97.33 & 20.14 & 41.48 & 72.71 & 85.80 & 74.83 & 85.17 & 62.96 & 77.44 \\
 & GRPO & 84.01 & 96.87 & 19.99 & 43.00 & 73.15 & 86.07 & 74.83 & 85.17 & 63.00 & 77.78 \\
 & SDPO & 83.97 & 97.27 & 19.77 & 42.75 & 72.98 & 86.07 & 74.76 & 84.75 & 62.87 & 77.71 \\
 & SCoRe & 84.45 & 96.13 & 20.42 & 41.98 & 72.16 & 85.89 & 76.35 & 90.58 & 63.34 & 78.65 \\
 & SEED & 84.04 & 97.33 & 19.72 & 43.77 & 72.96 & 85.89 & 74.85 & 84.83 & 62.89 & 77.96 \\
 & \ours{} & 87.30 & 96.73 & 29.89 & 47.07 & 74.50 & 86.07 & 80.32 & 93.50 & 68.00 & 80.84 \\
\midrule
\multirow{6}{*}{Olmo-3-7B-Instruct}
 & Random Sampling & 88.77 & 97.73 & 20.94 & 43.26 & 68.27 & 85.10 & 68.74 & 86.50 & 61.68 & 78.15 \\
 & GRPO & 88.79 & 97.47 & 20.93 & 43.00 & 68.13 & 85.19 & 68.38 & 86.33 & 61.56 & 78.00 \\
 & SDPO & 88.66 & 97.80 & 20.94 & 43.77 & 68.23 & 85.10 & 68.72 & 86.58 & 61.64 & 78.31 \\
 & SCoRe & 88.64 & 97.67 & 24.27 & 49.36 & 69.71 & 85.63 & 73.58 & 98.25 & 64.05 & 82.73 \\
 & SEED & 88.15 & 97.87 & 20.92 & 41.73 & 68.09 & 84.22 & 68.99 & 86.08 & 61.54 & 77.47 \\
 & \ours{} & 93.44 & 97.40 & 31.97 & 55.22 & 75.74 & 87.30 & 75.65 & 98.08 & 69.20 & 84.50 \\
\bottomrule
\end{tabular}}
\caption{Exploration results by benchmark, averaged over three run seeds (\%). Acc. is the average accuracy over all answers up to the cutoff step, and Pass is the cumulative pass rate at that step. Overall is the macro average over the four benchmarks. The cutoff step of a run is the last step reached by all four benchmarks, as in Table~\ref{tab:explore}. It is step 16 for SEED on Olmo-3-7B-Instruct at seeds 1 and 2 and step 24 otherwise.}
\label{tab:explore-bench}
\end{table*}

\paragraph{Seeds.} Each method is run with three seeds. Run $s\in\{1,2,3\}$ shifts every seed by $(s-1)\times10^{8}$, including the vLLM per-request decoding seed, the problem schedule of the batch methods, and the SEED analyzer seed. The problem set and the public/private split are the same for all three runs.

\subsection{Implementation details}

\paragraph{Shared configuration.} All batch baselines are implemented in verl. Each trains LoRA adapters with rank 64, alpha 64, and dropout 0 on all linear layers. We train in bfloat16 with FSDP, gradient checkpointing, and dynamic batch sizes. We use the AdamW optimizer with $\beta=(0.9,0.999)$, weight decay 0.01, a maximum gradient norm of 1.0, and a constant learning rate of $10^{-6}$ without warmup. Each update uses 32 questions, one PPO mini-batch of 32, and one epoch. Rollouts use the decoding settings of the main text: temperature 1.0 and top-$p$ 1.0 for Qwen3-8B, and temperature 0.6 and top-$p$ 0.95 for Olmo-3-7B-Instruct. The response limit is 8,192 tokens, except 16,384 tokens for Olmo-3-7B-Instruct on LCB. The batch methods apply a per-token rollout importance sampling correction with threshold 2.0. To spend $K=24$ verifier calls per problem with groups of 8 answers, each batch method runs $\lceil\lceil 24|\mathcal{D}|/8\rceil/32\rceil$ updates: 47 on MATH-500, 13 on LCB, 36 on MBPP+, and 38 on SciKnowEval. Table~\ref{tab:explore-hparams} summarizes the settings of each method.

\begin{table*}[t]
\centering
\small
\setlength{\tabcolsep}{4.5pt}
\begin{tabular}{llrrrrrr}
\toprule
 & & \multicolumn{3}{c}{Pass (\%) by seed} & \multicolumn{3}{c}{Mean over 3 seeds} \\
\cmidrule(lr){3-5}\cmidrule(lr){6-8}
Model & Method & Seed 1 & Seed 2 & Seed 3 & Pass (\%) & GPU h & PFLOPs \\
\midrule
\multirow{6}{*}{Qwen3-8B}
 & Random Sampling & 77.06 & 77.53 & 77.74 & $77.44\pm0.34$ & $0.66\pm0.03$ & $625\pm1$ \\
 & GRPO & 78.18 & 76.97 & 78.18 & $77.78\pm0.70$ & $4.43\pm0.09$ & $2{,}494\pm3$ \\
 & SDPO & 78.35 & 76.98 & 77.79 & $77.71\pm0.69$ & $5.06\pm0.10$ & $3{,}277\pm5$ \\
 & SCoRe & 78.52 & 78.85 & 78.57 & $78.65\pm0.18$ & $5.09\pm0.06$ & $3{,}058\pm5$ \\
 & SEED & 77.87 & 77.73 & 78.26 & $77.96\pm0.27$ & $7.87\pm0.16$ & $4{,}580\pm8$ \\
 & \ours{} & 80.94 & 81.00 & 80.59 & $80.84\pm0.22$ & $1.47\pm0.04$ & $1{,}043\pm2$ \\
\midrule
\multirow{6}{*}{Olmo-3-7B-Instruct}
 & Random Sampling & 77.81 & 78.54 & 78.09 & $78.15\pm0.37$ & $3.42\pm0.05$ & $676\pm3$ \\
 & GRPO & 77.24 & 78.70 & 78.05 & $78.00\pm0.73$ & $9.58\pm0.30$ & $2{,}885\pm34$ \\
 & SDPO & 78.65 & 77.89 & 78.39 & $78.31\pm0.38$ & $10.46\pm0.14$ & $3{,}843\pm26$ \\
 & SCoRe & 82.21 & 82.88 & 83.09 & $82.73\pm0.46$ & $12.33\pm0.02$ & $4{,}401\pm32$ \\
 & SEED & 76.78$^{\dagger}$ & 77.42$^{\dagger}$ & 78.23 & $77.47\pm0.73$ & $9.43\pm1.97$ & $3{,}796\pm1{,}152$ \\
 & \ours{} & 84.49 & 84.60 & 84.42 & $84.50\pm0.09$ & $6.43\pm0.12$ & $1{,}275\pm1$ \\
\bottomrule
\end{tabular}
\caption{Exploration pass rate (\%) by run seed, macro-averaged over MATH-500, LCB, MBPP+, and SciKnowEval, with the summed exploration cost underlying Table~\ref{tab:explore}. The last three columns are means $\pm$ sample standard deviations over 3 seeds. $^{\dagger}$The MATH-500 cell stopped at 16 steps, so the values are taken at step 16.}
\label{tab:explore-seeds}
\end{table*}

\begin{table*}[t]
\centering
\small
\setlength{\tabcolsep}{4.5pt}
\resizebox{\textwidth}{!}{%
\begin{tabular}{llrrrrr}
\toprule
Model & Method & MATH-500 & LCB & MBPP+ & SciKnowEval & Total \\
\midrule
\multirow{6}{*}{Qwen3-8B}
 & Random Sampling & 0.30 / 199 / 12.1 & 0.09 / 73 / 4.4 & 0.10 / 88 / 5.4 & 0.17 / 265 / 16.2 & 0.66 / 625 / 38.1 \\
 & GRPO & 1.82 / 998 / 12.2 & 1.01 / 362 / 4.4 & 0.67 / 441 / 5.4 & 0.92 / 693 / 7.3 & 4.43 / 2,494 / 29.3 \\
 & SDPO & 2.04 / 1,351 / 12.0 & 1.10 / 477 / 4.4 & 0.84 / 584 / 5.4 & 1.08 / 865 / 7.4 & 5.06 / 3,277 / 29.2 \\
 & SCoRe & 2.07 / 1,152 / 10.8 & 1.03 / 413 / 3.8 & 0.85 / 570 / 5.3 & 1.14 / 923 / 6.6 & 5.09 / 3,058 / 26.4 \\
 & SEED & 3.20 / 1,889 / 15.0 & 1.28 / 665 / 5.4 & 1.53 / 869 / 6.7 & 1.88 / 1,157 / 8.7 & 7.87 / 4,580 / 35.7 \\
 & \ours{} & 0.48 / 350 / 21.4 & 0.46 / 140 / 8.5 & 0.22 / 172 / 10.5 & 0.31 / 381 / 23.3 & 1.47 / 1,043 / 63.7 \\
\midrule
\multirow{6}{*}{Olmo-3-7B-Instruct}
 & Random Sampling & 1.35 / 227 / 15.5 & 1.45 / 157 / 10.8 & 0.38 / 90 / 6.2 & 0.23 / 201 / 13.8 & 3.42 / 676 / 46.3 \\
 & GRPO & 3.71 / 1,134 / 15.5 & 3.03 / 780 / 10.7 & 1.69 / 449 / 6.2 & 1.15 / 521 / 6.2 & 9.58 / 2,885 / 38.6 \\
 & SDPO & 4.12 / 1,562 / 15.5 & 3.30 / 1,057 / 10.6 & 1.75 / 596 / 6.1 & 1.28 / 628 / 6.2 & 10.46 / 3,843 / 38.4 \\
 & SCoRe & 4.52 / 1,612 / 16.9 & 3.92 / 1,085 / 11.4 & 2.31 / 862 / 9.0 & 1.58 / 842 / 6.6 & 12.33 / 4,401 / 43.9 \\
 & SEED & 3.32 / 1,506 / 15.1 & 2.76 / 1,006 / 10.2 & 1.81 / 629 / 6.2 & 1.53 / 656 / 6.0 & 9.43 / 3,796 / 37.5 \\
 & \ours{} & 2.16 / 450 / 30.8 & 2.80 / 296 / 20.3 & 1.01 / 245 / 16.7 & 0.47 / 284 / 19.4 & 6.43 / 1,275 / 87.3 \\
\bottomrule
\end{tabular}}
\caption{Exploration cost by benchmark, averaged over three run seeds. Each cell lists GPU h / PFLOPs / generated tokens (M), accumulated up to the cutoff step defined in Table~\ref{tab:explore-bench}. Total is the sum over the four benchmarks.}
\label{tab:explore-cost}
\end{table*}

\paragraph{Random Sampling.} Random Sampling draws one response per problem per step from the plain prompt with vLLM and does not update parameters. It uses the same decoding settings, response limits, and verifiers as the other methods.

\paragraph{GRPO.} GRPO samples 8 responses per question and normalizes advantages by the group standard deviation. It uses the vanilla clipped policy loss with a lower clip ratio of 0.2 and the higher clip bound of 0.28 from DAPO, without a KL term.

\paragraph{SDPO.} SDPO samples 8 responses per question and distills from a teacher whose prompt adds a successful response from the same group and the environment feedback on the failed attempt. A response is not used as the demonstration for itself. The teacher prompt is:
\begin{quote}\small\ttfamily
\{prompt\}\{solution\}\{feedback\}\par
\medskip
Correctly solve the original question.
\end{quote}
Distillation uses the top-20 tokens with a tail bucket, $\alpha=1.0$, an importance-sampling clip of 2.0, and an EMA teacher with update rate 0.01. The teacher prompt is limited to 10,240 tokens.

\paragraph{SCoRe.} SCoRe samples 4 two-turn trajectories per question, and the second turn uses the blind correction instruction of the task type. It uses a KL loss coefficient of 1.0 (\texttt{low\_var\_kl}), a clip ratio of 0.2, $\beta_1=0.01$, and $\beta_2=0.25$. The first 25\% of updates, rounded down, form stage 1 (11, 3, 9, and 9 updates on MATH-500, LCB, MBPP+, and SciKnowEval), and stage 2 adds a correction bonus of 10.0. Each turn is limited to 8,192 tokens, and the two turns together are limited to 16,896 tokens.

\paragraph{SEED.} SEED reserves 16\% of the updates, rounded to the nearest integer, for collecting hindsight skills: 8, 2, 6, and 6 updates' worth of rollouts on MATH-500, LCB, MBPP+, and SciKnowEval. These collection rollouts count toward the 24 verifier calls per problem. An analyzer writes skills at temperature 0, with feedback limited to 2,048 tokens and skills to 500 tokens. SFT on the collected data runs for 3 epochs with batch size 8. RL then runs for the remaining 39, 11, 30, and 32 updates with the SEED loss, a gate scale of 5.0, an on-policy distillation coefficient of 0.01, a reference KL coefficient of 0.01 (\texttt{low\_var\_kl}), and a clip ratio of 0.2. For Olmo-3-7B-Instruct, the token budget per RL update is limited to its maximum position embedding length of 65,536.

\paragraph{\ours{}.} \ours{} runs the rollout part of Algorithm~\ref{alg:ours} without parameter updates, using a restart interval of $M=5$ (restarts at steps 1, 6, 11, 16, and 21) and a tail length of $L=2{,}000$ tokens. Responses are generated with vLLM using GPU memory utilization 0.9, at most 16,384 batched tokens, and with prefix caching and chunked prefill enabled. The decoding settings, response limits, verifiers, and judge are the same as for Random Sampling.

\begin{table}[t]
\centering
\small
\setlength{\tabcolsep}{3.5pt}
\resizebox{\columnwidth}{!}{%
\begin{tabular}{lrrrrr}
\toprule
$x$-axis & Seed 1 & Seed 2 & Seed 3 & Mean & Mean curve \\
\midrule
Verifier calls & 1.97 & 2.20 & 1.63 & $1.94\pm0.29$ & 1.90 \\
PFLOPs & 6.19 & 6.91 & 5.25 & $6.12\pm0.83$ & 5.99 \\
GPU time & 3.85 & 4.10 & 3.33 & $3.76\pm0.40$ & 3.70 \\
\bottomrule
\end{tabular}}
\caption{Speedup of \ours{} over SCoRe ($\times$) on the two-model average curves of Figure~\ref{fig:explore}. We report the mean $\pm$ standard deviation of the per-seed ratios over 3 seeds. Mean curve is the ratio on the curve averaged over the three seeds.}
\label{tab:explore-speedup}
\end{table}

\section{Exploration results by benchmark and seed}
\label{app:explore-results}

This section reports the per-seed and per-benchmark measurements underlying the aggregate exploration results in Table~\ref{tab:explore} and Figure~\ref{fig:explore}.

\subsection{Results by seed}

Table~\ref{tab:explore-seeds} lists the per-seed pass rates whose means appear in Table~\ref{tab:explore}. On both models, \ours{} has the highest pass rate and SCoRe the second highest at every run seed. Across run seeds, the pass rate of \ours{} varies by less than 0.5 pp on each model.

\paragraph{Speedup over SCoRe.}
The speedup in Figure~\ref{fig:explore} is computed on the curve averaged over the two models. It is the ratio between the $x$ value of the last SCoRe point and the $x$ value at which \ours{} first reaches the same pass rate. Table~\ref{tab:explore-speedup} reports this ratio for each $x$-axis of Figure~\ref{fig:explore}. Against Random Sampling, the same definition on the verifier-call axis gives $4.07\times$. 

\subsection{Results by benchmark}

Table~\ref{tab:explore-bench} reports two values for each benchmark, averaged over the three run seeds. The first is the average accuracy, which is the fraction of correct answers among all graded answers up to the cutoff step. The second is the cumulative pass rate at that step. Overall is the macro average over the four benchmarks.

\begin{table*}[t]
\centering
\small
\setlength{\tabcolsep}{4.5pt}
\begin{tabular}{l r r r r r}
\toprule
 & & \multicolumn{2}{c}{Qwen3-8B} & \multicolumn{2}{c}{Olmo-3-7B-Instruct} \\
\cmidrule(lr){3-4}\cmidrule(lr){5-6}
Training set & Problems & $|\{x:\mathcal{C}^{+}(x)\neq\emptyset\}|$ & $(c^{+},y^{+})$ pairs & $|\{x:\mathcal{C}^{+}(x)\neq\emptyset\}|$ & $(c^{+},y^{+})$ pairs \\
\midrule
LCB (v2--v5) & 478 & 303.0 & 3,965 & 335.3 & 3,921 \\
MATH (Level 5) & 500 & 432.0 & 6,895 & 467.3 & 8,567 \\
SciKnowEval (L5) & 400 & 335.3 & 3,694 & 377.3 & 4,407 \\
\bottomrule
\end{tabular}
\caption{Training sets for in-distribution generalization and the data collected by the $K=20$ step exploration rollout of \ours{}, averaged over three run seeds. $|\{x:\mathcal{C}^{+}(x)\neq\emptyset\}|$ is the number of problems with at least one successful context, which is also the number of problems used in the policy update. $(c^{+},y^{+})$ pairs counts the distinct correct, untruncated answers with their contexts, summed over problems.}
\label{tab:general-train-data}
\end{table*}

On Qwen3-8B, \ours{} has the highest average accuracy on all four benchmarks (Table~\ref{tab:explore-bench}). Its pass rate is the highest on LCB, more than 3 pp above the next method, and on SciKnowEval, and ties with GRPO and SDPO for the highest on MBPP+. On MATH-500, its pass rate is above that of SCoRe and below those of the other methods. On Olmo-3-7B-Instruct, \ours{} again has the highest average accuracy on every benchmark (Table~\ref{tab:explore-bench}).

\subsection{Cost by benchmark}

Table~\ref{tab:explore-cost} reports the exploration cost of each benchmark, accumulated up to the cutoff step of the run and averaged over the three run seeds. GPU hours, PFLOPs, and generated tokens follow the accounting in Appendix~\ref{app:explore-setup}. 

On both models, \ours{} uses fewer GPU hours than each of the four RL baselines on every benchmark except Olmo-3-7B-Instruct LCB, where SEED uses slightly fewer, and more than Random Sampling on every benchmark (Table~\ref{tab:explore-cost}). On Olmo-3-7B-Instruct, its total GPU hours are about two thirds of those of the cheapest RL baseline, and LCB is its most expensive benchmark. The PFLOPs of \ours{} are below those of the four RL baselines on every benchmark for both models. This is consistent with the absence of parameter updates during exploration.

\section{Generalization setup details}
\label{app:general-setup}

This section specifies the training sets, test protocol, SFT baseline, training budgets and cost accounting underlying the in-distribution generalization results in Table~\ref{tab:general}. Unless stated otherwise, every method trains a separate LoRA adapter on each of the three training sets and is tested on the matching benchmark, and each run uses one H200 GPU. Table~\ref{tab:general} reports means over three run seeds. A run seed $s$ shifts the problem order, the decoding seed, the context draw of \ours{}, the analyzer seed of SEED, the rewrite seed of the SFT baseline and the test sampling seed by $(s-1)\times10^{8}$. The dataset files and the problem selection seed are fixed, so all three seeds use the same training and test problems. 

\subsection{Training sets}

The three training sets are disjoint from the test sets. For LCB, we take the problems in releases v2 to v5 of LiveCodeBench, that is, release v5 without the problems of release v1. We then apply static checks and remove duplicates as well as problems that collide with the 131 test problems, which leaves 478 problems. During training, all test cases of a training problem are used for reward and feedback. On these training sets, the code executor runs the tests in order and stops at the first failure, so the feedback shows a single failing test. For MATH, we sample 500 Level~5 problems from the MATH training split and drop problems whose text matches a MATH-500 problem. For SciKnowEval, we sample 100 open L5 problems from each of the biology, chemistry, materials and physics domains, and none of these 400 problems is in the 400-problem test set.

Table~\ref{tab:general-train-data} lists the training set sizes and the data collected by the exploration rollout of \ours{}. A problem enters the policy update only if its rollout produced at least one correct, untruncated answer, i.e., if $\mathcal{C}^{+}(x)\neq\emptyset$. The SFT baseline is built from the same $(c^{+},y^{+})$ pairs.

The policy update uses most of the problems in every training set, from about 63\% on LCB for Qwen3-8B to about 94\% on SciKnowEval for Olmo-3-7B-Instruct. LCB has the smallest fraction for both models, and Olmo-3-7B-Instruct uses a larger fraction than Qwen3-8B on every training set. MATH gives the most pairs for both models.

\subsection{Test protocol}

Each trained adapter is tested on its matching benchmark: LCB v6 (131 problems), MATH-500 (500 problems) or SciKnowEval (400 problems). We sample four responses per problem with the decoding settings and response length limit of training, using a separate seed for each response. LCB responses are graded on all test cases, both public and private. MATH-500 responses are graded with math-verify on the last \texttt{\textbackslash boxed\{\}}, and SciKnowEval responses with the base-model judge. Let $z_{i,j}\in\{0,1\}$ indicate whether response $j$ to problem $i$ is correct. With $P$ test problems, we compute avg@4 and pass@4. Overall is the arithmetic mean of avg@4 on LCB, MATH-500 and SciKnowEval.

\begin{table*}[t]
\centering
\small
\setlength{\tabcolsep}{3.5pt}
\begin{tabular}{l l r r r l r r r}
\toprule
Model & Training set & Pairs & Kept & Dropped & Kept per round (1/2/3/4/5) & Length & Trunc. (\%) & GPU h \\
\midrule
\multirow{3}{*}{Qwen3-8B} & LCB & 3,965 & 3,959 & 6 & 3,940 / 18 / 1 / 0 / 0 & 349 & 0.00 & 0.44 \\
 & MATH & 6,895 & 6,895 & 0 & 6,889 / 5 / 1 / 0 / 0 & 737 & 0.01 & 0.37 \\
 & SciKnowEval & 3,694 & 3,588 & 106 & 3,415 / 96 / 41 / 22 / 15 & 1,031 & 0.03 & 0.39 \\
\midrule
\multirow{3}{*}{Olmo-3-7B-Instruct} & LCB & 3,921 & 3,899 & 22 & 3,823 / 57 / 12 / 5 / 2 & 404 & 0.00 & 0.58 \\
 & MATH & 8,567 & 8,566 & 1 & 8,562 / 2 / 0 / 1 / 0 & 597 & 0.00 & 0.95 \\
 & SciKnowEval & 4,407 & 3,964 & 443 & 3,031 / 484 / 227 / 136 / 86 & 899 & 3.96 & 1.49 \\
\bottomrule
\end{tabular}
\caption{Statistics of the rewritten solutions $y^{*}$ used by the SFT baseline, averaged over three run seeds and rounded. Pairs is the number of $(c^{+},y^{+})$ pairs. Kept and Dropped count pairs whose rewrite did or did not pass the verifier within 5 rounds. Kept per round gives the number of pairs first accepted in each round. Length is the mean number of tokens of the kept rewrites. Trunc. is the fraction of truncated rewrite generations over all rounds. GPU h covers rewriting and verification.}
\label{tab:sft-rewrite-stats}
\end{table*}

\subsection{SFT baseline}

The SFT baseline tests whether training on the correct answers collected by the rollout of \ours{} gives the same gain as self-distillation. It uses the same rollout as \ours{} and runs three stages: rewriting, verification and supervised training.

\paragraph{Rewriting.}
For every pair of a successful context $c^{+}$ and its correct answer $y^{+}$, the base model of the run rewrites $y^{+}$ into a self-contained solution $y^{*}$. The request consists of the system turn of the plain prompt, if there is one, and one user turn. When $c^{+}$ contains an earlier attempt, the user turn is:
\begin{quote}\small\ttfamily
\{plain user prompt\}\par
\medskip
Below is an earlier solving session for this problem: earlier attempts, the feedback on them, and a final solution that was verified to be correct.\par
\medskip
<session>\par
\{transcript\}\par
</session>\par
\medskip
<correct\_solution>\par
\{y+\}\par
</correct\_solution>\par
\medskip
Rewrite the correct solution into a complete, self-contained solution to the problem, written as if you were solving it for the first time. Remove the wrong attempts, the invalid derivations, the wrong intermediate results and every reference to earlier attempts, feedback, checking or corrections. Present the correct reasoning in a logical order, keeping its intermediate results. \{ending\} Output only the rewritten solution.
\end{quote}
The transcript lists the turns of $c^{+}$ after the plain prompt, labeled \texttt{[Problem prompt]}, \texttt{[Earlier attempt]} and \texttt{[Feedback]}. When $c^{+}$ has no earlier attempt, the sentence and the session block are replaced by ``\texttt{Below is a solution to this problem that was verified to be correct.}'' The ending keeps the final answer of $y^{+}$ fixed. For MATH it is:
\begin{quote}\small\ttfamily
The final answer must be exactly the same as in the correct solution: end with \textbackslash boxed\{\{answer\}\}.
\end{quote}
Here, \texttt{\{answer\}} is the last boxed answer of $y^{+}$. For LCB, the ending requires the same complete program in a single \texttt{python} code block. For SciKnowEval, it requires exactly the same final answer after ``\texttt{Final answer:}''. Rewrites are sampled at temperature 0.7 and top-$p$ 0.95, with the response length limit of training.

\begin{table*}[t]
\centering
\small
\setlength{\tabcolsep}{3.2pt}
\resizebox{\textwidth}{!}{%
\begin{tabular}{l l r r r r}
\toprule
Model & Method & LCB & MATH & SciKE & Overall \\
\midrule
\multirow{7}{*}{Qwen3-8B}
 & Base & 19.08 / 18.89 / 21.37 & 84.45 / 83.60 / 84.20 & 74.69 / 74.00 / 74.50 & 59.41 / 58.83 / 60.02 \\
 & GRPO & 25.38 / 25.95 / 26.72 & 86.35 / 86.90 / 86.35 & 75.62 / 77.25 / 77.38 & 62.45 / 63.37 / 63.48 \\
 & SDPO & 25.57 / 25.57 / 25.76 & 82.00 / 81.40 / 81.55 & 73.56 / 72.12 / 73.19 & 60.38 / 59.70 / 60.17 \\
 & SCoRe & 22.52 / 22.52 / 23.09 & 84.20 / 84.70 / 84.05 & 75.31 / 75.06 / 75.12 & 60.68 / 60.76 / 60.76 \\
 & SEED & 26.34 / 25.00 / 26.34 & 86.20 / 84.55 / 86.00 & 76.75 / 77.19 / 76.56 & 63.10 / 62.25 / 62.97 \\
 & SFT & 22.52 / 23.47 / 20.80 & 77.40 / 77.00 / 76.15 & 76.00 / 75.50 / 74.62 & 58.64 / 58.66 / 57.19 \\
 & \ours{} & 26.15 / 25.76 / 24.43 & 91.75 / 92.15 / 91.65 & 76.88 / 77.25 / 76.19 & 64.92 / 65.05 / 64.09 \\
\midrule
\multirow{7}{*}{Olmo-3-7B-Instruct}
 & Base & 21.56 / 22.14 / 20.04 & 89.15 / 89.35 / 88.85 & 68.81 / 69.19 / 69.75 & 59.84 / 60.22 / 59.55 \\
 & GRPO & 21.76 / 22.14 / 19.47 & 90.10 / 88.80 / 89.10 & 69.56 / 72.38 / 70.56 & 60.47 / 61.10 / 59.71 \\
 & SDPO & 22.71 / 24.24 / 24.05 & 82.00 / 83.60 / 81.55 & 66.88 / 67.81 / 66.25 & 57.19 / 58.55 / 57.28 \\
 & SCoRe & 20.61 / 20.42 / 20.42 & 89.05 / 88.95 / 88.85 & 68.62 / 68.56 / 67.94 & 59.43 / 59.31 / 59.07 \\
 & SEED & 21.76 / 22.52 / 22.71 & 89.65 / 88.70 / 88.95 & 71.81 / 73.12 / 71.38 & 61.07 / 61.45 / 61.01 \\
 & SFT & 23.47 / 22.14 / 22.71 & 81.95 / 80.95 / 81.10 & 70.25 / 71.06 / 70.25 & 58.56 / 58.05 / 58.02 \\
 & \ours{} & 24.05 / 23.28 / 24.81 & 90.40 / 90.45 / 91.00 & 73.69 / 75.00 / 74.38 & 62.71 / 62.91 / 63.39 \\
\bottomrule
\end{tabular}}
\caption{Test avg@4 results by run seed (\%). Values are ordered by run seed 1 / 2 / 3. Overall is the mean of LCB, MATH and SciKE for the same run seed.}
\label{tab:general-seed}
\end{table*}

\paragraph{Verification.}
Each rewrite is checked by the verifier at training time: all test cases of the training problem for LCB, math-verify for MATH, and the LLM judge for SciKnowEval. A rewrite is kept only if it is judged correct. A pair whose rewrites fail in all 5 rounds is dropped.

\paragraph{Supervised fine-tuning.}
The SFT model learns to map the plain prompt $x$ to $y^{*}$ with a negative log-likelihood loss. It uses the same LoRA configuration, learning rate ($10^{-5}$), batch size and number of steps as the policy update of \ours{}: 128 examples per step for 20 steps, i.e., 2,560 examples per run. Every training set in Table~\ref{tab:sft-rewrite-stats} has more kept rewrites than this, so each run uses less than one epoch.

Table~\ref{tab:sft-rewrite-stats} reports the rewrite statistics, averaged over the three run seeds.

On LCB and MATH, at most 22 pairs per training set are dropped. SciKnowEval has the most dropped pairs and the longest kept rewrites, 1,031 and 899 tokens on average. Supervised training alone takes 0.55 and 0.52 GPU hours.

\begin{table*}[t]
\centering
\small
\setlength{\tabcolsep}{3.5pt}
\begin{tabular}{l l r r r r}
\toprule
Model & Method & LCB & MATH & SciKE & Overall \\
\midrule
\multirow{7}{*}{Qwen3-8B}
 & Base & 19.78 / 31.55 & 84.08 / 93.53 & 74.40 / 81.00 & 59.42 \\
 & GRPO & 26.02 / 35.62 & 86.53 / 94.33 & 76.75 / 82.33 & 63.10 \\
 & SDPO & 25.64 / 34.10 & 81.65 / 91.40 & 72.96 / 80.58 & 60.08 \\
 & SCoRe & 22.71 / 32.82 & 84.32 / 93.60 & 75.17 / 81.92 & 60.73 \\
 & SEED & 25.89 / 35.88 & 85.58 / 93.87 & 76.83 / 82.42 & 62.77 \\
 & SFT & 22.26 / 33.08 & 76.85 / 88.87 & 75.38 / 84.83 & 58.16 \\
 & \ours{} & 25.45 / 34.86 & 91.85 / 96.20 & 76.77 / 82.25 & 64.69 \\
\midrule
\multirow{7}{*}{Olmo-3-7B-Instruct}
 & Base & 21.25 / 29.26 & 89.12 / 95.80 & 69.25 / 80.50 & 59.87 \\
 & GRPO & 21.12 / 30.79 & 89.33 / 96.00 & 70.83 / 82.67 & 60.43 \\
 & SDPO & 23.66 / 34.10 & 82.38 / 91.60 & 66.98 / 78.58 & 57.68 \\
 & SCoRe & 20.48 / 29.77 & 88.95 / 95.53 & 68.38 / 79.75 & 59.27 \\
 & SEED & 22.33 / 33.59 & 89.10 / 95.60 & 72.10 / 82.50 & 61.18 \\
 & SFT & 22.77 / 32.82 & 81.33 / 91.67 & 70.52 / 83.00 & 58.21 \\
 & \ours{} & 24.05 / 36.13 & 90.62 / 96.13 & 74.35 / 86.25 & 63.01 \\
\bottomrule
\end{tabular}
\caption{Test avg@4 / pass@4 (\%), averaged over three run seeds. Overall is the mean avg@4 over LCB, MATH and SciKE.}
\label{tab:general-pass4}
\end{table*}

\subsection{Training budgets}

\paragraph{Shared training configuration.}
All trained methods use LoRA adapters with the configuration of the exploration experiments (Appendix~\ref{app:explore-setup}) and a constant learning rate of $10^{-5}$. Training rollouts use the decoding settings of the exploration experiments: temperature 1.0 and top-$p$ 1.0 for Qwen3-8B with thinking disabled, and temperature 0.6 and top-$p$ 0.95 for Olmo-3-7B-Instruct. The maximum response length is 8,192 tokens, except for Olmo-3-7B-Instruct on LCB, where it is 16,384 tokens.

\paragraph{GRPO, SDPO and SCoRe.}
Each of these baselines receives a budget of 20 verifier calls per training problem. They sample 8 responses per problem (for SCoRe, 4 trajectories of 2 turns each) and use 32 problems per update, so the budget gives 38, 40 and 32 updates on LCB, MATH and SciKnowEval, respectively. The remaining settings are the same as in exploration.

\paragraph{SEED.}
At a budget of 20 verifier calls per problem, the seed-1 SEED training took 9.58 GPU hours on Qwen3-8B and 19.07 on Olmo-3-7B-Instruct, compared with 5.32--6.00 and 14.03--17.65 for GRPO, SDPO and SCoRe. We therefore lower its budget to 15 verifier calls per training problem. This gives 29, 30 and 24 updates on LCB, MATH and SciKnowEval, of which the first 5, 5 and 4 form the collection stage. SFT and RL follow as in exploration.

\paragraph{\ours{}.}
\ours{} first runs the rollout part of Algorithm~\ref{alg:ours} on each training set for $K=20$ steps. It uses $M=5$, $L=2{,}000$ and the feedback of the exploration experiments, so the MATH feedback contains only the verdict. The policy update then runs $S=20$ steps with a batch of 128 problems and one optimizer step per training step. For each problem in the batch, the update samples one student response from the plain prompt and draws a context uniformly from $\mathcal{C}^{+}(x)$, separately for each step and problem. The divergence is the reverse KL in Equation~\ref{eq:fdiv}, with $\tau=10^{-3}$ on MATH and no cap on LCB and SciKnowEval, averaged over tokens as in Equation~\ref{eq:loss}.

\subsection{Cost accounting}

GPU hours in Table~\ref{tab:general} are the summed elapsed times of a run's completed or timed-out jobs on the three training sets. FLOPs use the estimate of Appendix~\ref{app:explore-setup}: $2N$ per forward token plus $6N$ per trained token. For \ours{}, the cost includes the $K=20$ step exploration rollout and the policy update. In the update, forward tokens cover student generation and the teacher and behavior forward passes, and trained tokens are the student tokens. For SFT, the cost includes the same rollout, the rewriting and verification in Table~\ref{tab:sft-rewrite-stats}, and supervised training. The cost of SEED includes its collection stage. Averaged over the three run seeds, the exploration rollout takes 1.58 GPU hours for Qwen3-8B and 9.93 for Olmo-3-7B-Instruct, and the policy update of \ours{} takes 2.12 and 4.11 GPU hours.

\begin{table*}[t]
\centering
\small
\setlength{\tabcolsep}{3pt}
\resizebox{\textwidth}{!}{%
\begin{tabular}{l l r r r r r r r r r r r}
\toprule
 & & \multicolumn{7}{c}{GPU h} & \multicolumn{4}{c}{EFLOPs} \\
\cmidrule(lr){3-9}\cmidrule(lr){10-13}
Model & Method & LCB & MATH & SciKE & Train & Rollout & Rewrite & Total & Train & Rollout & Rewrite & Total \\
\midrule
\multirow{6}{*}{Qwen3-8B}
 & GRPO & 1.78 & 2.42 & 1.31 & 5.52 & -- & -- & 5.52 & 3.71 & -- & -- & 3.71 \\
 & SDPO & 1.84 & 2.03 & 1.96 & 5.82 & -- & -- & 5.82 & 4.35 & -- & -- & 4.35 \\
 & SCoRe & 1.69 & 2.40 & 1.47 & 5.57 & -- & -- & 5.57 & 4.17 & -- & -- & 4.17 \\
 & SEED & 2.21 & 2.79 & 1.71 & 6.72 & -- & -- & 6.72 & 4.88 & -- & -- & 4.88 \\
 & SFT & 0.16 & 0.17 & 0.22 & 0.55 & 1.58 & 1.19 & 3.32 & 0.39 & 1.36 & 0.69 & 2.44 \\
 & \ours{} & 0.49 & 0.98 & 0.65 & 2.12 & 1.58 & -- & 3.70 & 1.28 & 1.36 & -- & 2.64 \\
\midrule
\multirow{6}{*}{Olmo-3-7B-Instruct}
 & GRPO & 6.91 & 5.46 & 1.70 & 14.06 & -- & -- & 14.06 & 4.52 & -- & -- & 4.52 \\
 & SDPO & 8.59 & 4.11 & 1.53 & 14.23 & -- & -- & 14.23 & 5.93 & -- & -- & 5.93 \\
 & SCoRe & 9.30 & 6.27 & 2.15 & 17.71 & -- & -- & 17.71 & 6.33 & -- & -- & 6.33 \\
 & SEED & 6.72 & 5.40 & 2.44 & 14.56 & -- & -- & 14.56 & 5.94 & -- & -- & 5.94 \\
 & SFT & 0.17 & 0.14 & 0.21 & 0.52 & 9.93 & 3.02 & 13.47 & 0.33 & 1.70 & 0.97 & 3.00 \\
 & \ours{} & 1.90 & 1.46 & 0.75 & 4.11 & 9.93 & -- & 14.03 & 1.44 & 1.70 & -- & 3.15 \\
\bottomrule
\end{tabular}}
\caption{Training cost by training set and stage, averaged over three run seeds. LCB, MATH and SciKE give the training GPU h on each training set, and Train is their sum. Rollout is the exploration rollout that collects the training data of \ours{} and SFT, and Rewrite is the generation and verification of the SFT targets. Total GPU h and Total EFLOPs ($10^{18}$ FLOPs) equal the values in Table~\ref{tab:general}. -- marks a stage that the method does not use.}
\label{tab:general-cost}
\end{table*}

\section{Generalization results by benchmark and seed}
\label{app:general-results}

This section reports the measurements underlying the in-distribution generalization results in Table~\ref{tab:general}. Settings, training sets and cost accounting follow Appendix~\ref{app:general-setup}. Each trained method is run with three run seeds, 1, 2, and 3. The training and test problems are identical across run seeds. Base is not trained, so its values vary across run seeds only through the test-sampling seed. Test accuracy is avg@4 (\%) on LCB v6, MATH-500 and SciKnowEval, and Overall is the arithmetic mean of the three. Pass@4 and the cost breakdown by training set are also reported as means over the three run seeds.

\subsection{Results by run seed}

Table~\ref{tab:general-seed} reports the seed-level test avg@4 underlying the means in Table~\ref{tab:general}. Each cell lists run seeds 1 / 2 / 3 in that order. The seed-1 values of Base and \ours{} are the single-run values used in the ablations, for example in Table~\ref{tab:loss}. All three SEED run seeds use a budget of 15 verifier calls per training problem. Per-seed GPU h and EFLOPs are not reported separately.

On Qwen3-8B, \ours{} has the highest Overall at every run seed. The second-highest Overall is SEED at run seed 1 and GRPO at run seeds 2 and 3. \ours{} also has the highest MATH-500 accuracy at every run seed, more than 5 pp above GRPO, the second-highest method. On LCB, \ours{} is not the highest at any run seed, where SEED is highest at run seed 1 and GRPO at run seeds 2 and 3. SFT scores below Base in Overall and on MATH-500 at every run seed.

On Olmo-3-7B-Instruct, \ours{} has the highest Overall at every run seed, and SEED has the second highest at every run seed. \ours{} also has the highest MATH-500 and SciKnowEval accuracy at every run seed. On LCB, SDPO exceeds \ours{} at run seed 2. SFT scores below Base in Overall at every run seed, and SDPO's MATH-500 accuracy is more than 5 pp below Base at every run seed.

\subsection{Pass@4}

Table~\ref{tab:general-pass4} reports test avg@4 and pass@4, averaged over the three run seeds. Here, avg@4 is the fraction of all four responses per problem that are correct, and pass@4 is the fraction of test problems with at least one correct response. The Overall column equals the Overall values in Table~\ref{tab:general}.

\begin{table*}[t]
\centering
\small
\setlength{\tabcolsep}{4.5pt}
\begin{tabular}{l r r r r r r r r}
\toprule
 & \multicolumn{4}{c}{Qwen3-8B} & \multicolumn{4}{c}{Olmo-3-7B-Instruct} \\
\cmidrule(lr){2-5}\cmidrule(lr){6-9}
Method & MATH-500 & LCB & MBPP+ & SciKE & MATH-500 & LCB & MBPP+ & SciKE \\
\midrule
Random Sampling & 883 & 841 & 395 & 344 & 1,175 & 2,821 & 483 & 207 \\
GRPO & 886 & 839 & 395 & 345 & 1,175 & 2,792 & 478 & 206 \\
SDPO & 871 & 835 & 391 & 353 & 1,171 & 2,767 & 471 & 204 \\
SCoRe & 802 & 900 & 455 & 395 & 1,310 & 3,266 & 853 & 385 \\
SEED & 885 & 836 & 392 & 346 & 1,159 & 2,829 & 485 & 209 \\
\ours{} & 694 & 859 & 382 & 416 & 1,194 & 3,598 & 791 & 292 \\
\bottomrule
\end{tabular}
\caption{Mean response tokens per graded answer during exploration, averaged over three run seeds.}
\label{tab:len-explore-bench}
\end{table*}

\begin{table*}[t]
\centering
\small
\setlength{\tabcolsep}{3.5pt}
\begin{tabular}{l r r r r r r}
\toprule
 & \multicolumn{3}{c}{Qwen3-8B} & \multicolumn{3}{c}{Olmo-3-7B-Instruct} \\
\cmidrule(lr){2-4}\cmidrule(lr){5-7}
Method & LCB & MATH-500 & SciKE & LCB & MATH-500 & SciKE \\
\midrule
Base & 843 / 0.00 & 879 / 0.12 & 341 / 0.00 & 2,781 / 2.61 & 1,166 / 2.95 & 205 / 0.15 \\
GRPO & 936 / 0.00 & 1,001 / 0.35 & 412 / 0.00 & 3,049 / 2.54 & 1,201 / 3.10 & 254 / 0.29 \\
SDPO & 808 / 0.00 & 678 / 0.00 & 471 / 1.40 & 3,674 / 2.42 & 747 / 1.00 & 171 / 0.02 \\
SCoRe & 849 / 0.00 & 878 / 0.10 & 350 / 0.04 & 2,765 / 1.72 & 1,190 / 3.23 & 205 / 0.13 \\
SEED & 828 / 0.06 & 925 / 0.12 & 394 / 0.02 & 2,677 / 1.27 & 1,219 / 3.20 & 304 / 0.90 \\
SFT & 538 / 0.06 & 594 / 0.00 & 423 / 0.08 & 2,640 / 1.27 & 751 / 1.10 & 291 / 0.65 \\
\ours{} & 1,167 / 0.00 & 1,864 / 6.33 & 477 / 0.13 & 4,264 / 4.01 & 1,401 / 4.00 & 398 / 0.27 \\
\bottomrule
\end{tabular}
\caption{Test response length after the policy update. Each cell lists the mean completion tokens and the truncation rate (\%), both averaged over three run seeds. Means include truncated responses.}
\label{tab:len-general-bench}
\end{table*}

On Qwen3-8B, \ours{} has the highest pass@4 on MATH-500, about 2 pp above GRPO, the second highest method. On LCB, SEED and GRPO have a higher pass@4 than \ours{}, and SFT has the highest pass@4 on SciKnowEval. On Olmo-3-7B-Instruct, \ours{} has the highest pass@4 on all three test sets.

\subsection{Training cost by training set}

Table~\ref{tab:general-cost} separates the cost by training set and by stage, averaged over the three run seeds. GPU h and FLOPs follow the accounting in Appendix~\ref{app:general-setup}. For SEED, the training columns include its collection stage. For \ours{} and SFT, Rollout is the 20-step exploration rollout on the three training sets that produces the training data. For SFT, Rewrite is the generation and verification of the rewritten solutions. Its FLOPs count $2N$ per prompt and completion token of the rewriting calls and exclude the verification, with $N=8.19\times10^9$ for Qwen3-8B and $7.30\times10^9$ for Olmo-3-7B-Instruct.

For \ours{}, the exploration rollout accounts for less than half of the total GPU h on Qwen3-8B and about 70\% on Olmo-3-7B-Instruct, where the LCB rollout is the largest part. The training FLOPs of \ours{} are a quarter to a third of those of the four RL baselines on both models. On Qwen3-8B, the MATH training set has the largest training GPU h for every method except SFT, and on Olmo-3-7B-Instruct the LCB training set does. For SFT, the rewriting stage takes more GPU time than supervised training on both models.

\section{Additional ablation results}
\label{app:ablation}

This section gives the per-benchmark measurements behind the ablation studies in the main text, in the same order as the main text. Unless stated otherwise, every ablation is a single run with run seed 1. The policy-update ablations are compared with the seed-1 Base and \ours{} checkpoints. The response length analysis is the exception and uses all three run seeds. Exploration cells report ``mean accuracy over the 24 steps / cumulative pass rate at the last step''. On LCB and MBPP+, an answer counts as correct only if it passes both the public and the private tests. GPU hours, PFLOPs and generated tokens of exploration follow the accounting in Appendix~\ref{app:explore-setup}, and training costs of the policy update follow Appendix~\ref{app:general-setup}.

\begin{table*}[t]
\centering
\small
\setlength{\tabcolsep}{3.2pt}
\resizebox{\textwidth}{!}{%
\begin{tabular}{l l r r r r r}
\toprule
Model & Variant & MATH-500 & LCB & MBPP+ & SciKE & Total \\
\midrule
\multirow{7}{*}{Qwen3-8B} & \ours{} & 0.48 / 350 / 21.4 & 0.46 / 140 / 8.6 & 0.22 / 173 / 10.6 & 0.32 / 380 / 23.2 & 1.47 / 1,044 / 63.7 \\
 & w/o last response & 0.55 / 280 / 17.1 & 0.47 / 108 / 6.6 & 0.17 / 139 / 8.5 & 0.27 / 326 / 19.9 & 1.47 / 854 / 52.1 \\
 & w/o feedback & 0.52 / 371 / 22.6 & 0.49 / 140 / 8.5 & 0.20 / 194 / 11.8 & 0.35 / 434 / 26.5 & 1.56 / 1,139 / 69.5 \\
 & Restart every 2 & 0.63 / 421 / 25.7 & 0.46 / 164 / 10.0 & 0.25 / 213 / 13.0 & 0.33 / 419 / 25.6 & 1.68 / 1,217 / 74.3 \\
 & Restart every 10 & 0.43 / 331 / 20.2 & 0.50 / 134 / 8.2 & 0.21 / 160 / 9.8 & 0.29 / 367 / 22.4 & 1.43 / 992 / 60.6 \\
 & No restart & 0.39 / 314 / 19.2 & 0.52 / 124 / 7.5 & 0.20 / 145 / 8.9 & 0.25 / 351 / 21.4 & 1.36 / 934 / 57.0 \\
 & Last 3 rounds & 0.57 / 483 / 29.5 & 0.58 / 195 / 11.9 & 0.24 / 246 / 15.0 & 0.35 / 453 / 27.6 & 1.74 / 1,377 / 84.0 \\
\midrule
\multirow{7}{*}{Olmo-3-7B-Instruct} & \ours{} & 2.04 / 449 / 30.7 & 2.81 / 298 / 20.4 & 0.99 / 244 / 16.7 & 0.46 / 284 / 19.4 & 6.30 / 1,274 / 87.3 \\
 & w/o last response & 2.07 / 361 / 24.7 & 2.41 / 237 / 16.2 & 0.86 / 151 / 10.3 & 0.42 / 249 / 17.1 & 5.77 / 998 / 68.3 \\
 & w/o feedback & 2.98 / 563 / 38.5 & 2.24 / 264 / 18.1 & 1.42 / 331 / 22.7 & 0.56 / 324 / 22.2 & 7.19 / 1,482 / 101.5 \\
 & Restart every 2 & 2.85 / 560 / 38.4 & 3.44 / 354 / 24.2 & 1.36 / 312 / 21.4 & 0.53 / 321 / 22.0 & 8.18 / 1,548 / 106.0 \\
 & Restart every 10 & 1.94 / 414 / 28.3 & 2.61 / 273 / 18.7 & 0.87 / 222 / 15.2 & 0.43 / 272 / 18.6 & 5.85 / 1,181 / 80.9 \\
 & No restart & 1.61 / 386 / 26.4 & 2.17 / 250 / 17.1 & 0.74 / 190 / 13.0 & 0.45 / 258 / 17.6 & 4.97 / 1,083 / 74.2 \\
 & Last 3 rounds & 2.66 / 629 / 43.1 & 2.59 / 341 / 23.4 & 1.12 / 359 / 24.6 & 0.48 / 334 / 22.9 & 6.84 / 1,663 / 113.9 \\
\bottomrule
\end{tabular}}
\caption{Exploration cost of the component ablations. Each cell lists GPU hours, PFLOPs and generated tokens (millions), cumulative at the last step of the benchmark. Total is the sum over the four benchmarks.}
\label{tab:abl-components-cost}
\end{table*}

\subsection{Response length}
\label{app:abl-length}

Table~\ref{tab:len-explore-bench} reports the per-benchmark response lengths underlying Table~\ref{tab:length}(a), and Table~\ref{tab:len-general-bench} reports those underlying Table~\ref{tab:length}(b). For exploration, the length is the number of response tokens per graded answer. For the policy update, it is the mean number of completion tokens over all four test responses per problem, including truncated responses. Both are averaged over the three run seeds. A test response is truncated when its finish reason is \texttt{length} or it reaches the response limit. For each model and benchmark, the length ratio in Table~\ref{tab:length} divides the three-seed mean length of a method by that of the reference (Random Sampling or Base), and the gain subtracts the three-seed mean accuracy of the reference from that of the method. Table~\ref{tab:length} reports the arithmetic means of the ratio and the gain over the model and benchmark cells, eight in exploration and six in generalization.

During exploration, \ours{} writes longer answers than Random Sampling on LCB and SciKnowEval for both models and on Olmo-3-7B-Instruct MBPP+, with the largest relative difference on Olmo-3-7B-Instruct MBPP+. On MATH-500, its answers are shorter on Qwen3-8B and close to Random Sampling on Olmo-3-7B-Instruct. SCoRe is longer than Random Sampling in seven of the eight cells, and GRPO, SDPO and SEED stay within 3\% of Random Sampling in every cell. After the policy update, \ours{} has the longest test responses in all six model and test-set cells (Table~\ref{tab:len-general-bench}). The largest relative increase over Base is on Qwen3-8B MATH-500, where the mean length more than doubles and the truncation rate rises to 6.33\%. SFT shortens LCB and MATH-500 responses for both models.

\begin{table*}[t]
\centering
\small
\setlength{\tabcolsep}{3.5pt}
\begin{tabular}{l l r r r r r}
\toprule
Model & Variant & MATH-500 & LCB & MBPP+ & SciKE & Overall \\
\midrule
\multirow{7}{*}{Qwen3-8B} & \ours{} & 87.12 / 97.20 & 29.48 / 47.33 & 74.17 / 86.24 & 80.52 / 93.00 & 67.82 / 80.94 \\
 & w/o last response & 84.07 / 98.00 & 21.56 / 46.56 & 71.88 / 86.24 & 77.02 / 88.00 & 63.63 / 79.70 \\
 & w/o feedback & 85.92 / 95.60 & 22.30 / 38.93 & 71.10 / 83.60 & 80.67 / 93.25 & 65.00 / 77.84 \\
 & Restart every 2 & 86.15 / 97.60 & 24.71 / 46.56 & 72.72 / 86.77 & 79.30 / 92.50 & 65.72 / 80.86 \\
 & Restart every 10 & 87.75 / 95.80 & 31.71 / 47.33 & 75.18 / 84.92 & 81.52 / 94.50 & 69.04 / 80.64 \\
 & No restart & 88.47 / 95.40 & 36.04 / 47.33 & 76.17 / 83.33 & 82.41 / 93.25 & 70.77 / 79.83 \\
 & Last 3 rounds & 86.83 / 95.20 & 30.31 / 47.33 & 75.60 / 86.51 & 81.12 / 93.50 & 68.46 / 80.63 \\
\midrule
\multirow{7}{*}{Olmo-3-7B-Instruct} & \ours{} & 93.68 / 97.40 & 32.38 / 55.73 & 76.17 / 87.57 & 76.13 / 97.25 & 69.59 / 84.49 \\
 & w/o last response & 90.75 / 97.80 & 22.39 / 49.62 & 69.40 / 86.77 & 70.09 / 95.75 & 63.16 / 82.49 \\
 & w/o feedback & 92.86 / 97.60 & 29.23 / 53.44 & 73.41 / 83.86 & 81.03 / 100.00 & 69.13 / 83.72 \\
 & Restart every 2 & 92.78 / 97.80 & 30.38 / 54.96 & 74.34 / 89.15 & 77.49 / 99.50 & 68.75 / 85.35 \\
 & Restart every 10 & 94.25 / 97.40 & 34.03 / 57.25 & 76.87 / 88.36 & 75.54 / 96.75 & 70.17 / 84.94 \\
 & No restart & 95.32 / 97.80 & 35.18 / 55.73 & 77.50 / 84.66 & 73.72 / 97.00 & 70.43 / 83.80 \\
 & Last 3 rounds & 93.25 / 97.60 & 33.68 / 56.49 & 75.72 / 86.24 & 77.79 / 98.50 & 70.11 / 84.71 \\
\bottomrule
\end{tabular}
\caption{Component ablations during exploration (\%). Each cell lists the mean accuracy over the 24 steps and the cumulative pass rate at step 24. Overall is the unweighted mean over the four benchmarks. Single run with run seed 1.}
\label{tab:abl-components-acc}
\end{table*}

\subsection{Restart, last response, feedback and history}
\label{app:abl-components}

This section reports the component ablations in Figure~\ref{fig:ablation} separately for each benchmark. Each variant changes one part of the context of \ours{} and keeps the rest of the exploration setting. This includes the budget of $\min(24\text{ steps}, 5\text{ GPU h})$ per benchmark and the H200 engine parameters.
\begin{itemize}
\item \ours{} w/o last response replaces the dialogue with a single user turn. This turn holds the problem, the feedback on the unseen previous answer and the revision instruction.
\item \ours{} w/o feedback keeps the previous answer but replaces the feedback with the blind correction instruction. The verifier still runs for grading.
\item The restart variants restart every 2 or 10 steps instead of every 5, and the no-restart variant restarts only at step 1.
\item The last-3 variant keeps the last three rounds since the most recent restart. Each round holds a compressed response and its feedback, and only the newest user turn asks for a revision.
\end{itemize}
Table~\ref{tab:abl-components-acc} reports accuracy and pass rate, and Table~\ref{tab:abl-components-cost} reports the cost.

\begin{table*}[t]
\centering
\small
\setlength{\tabcolsep}{3pt}
\resizebox{\textwidth}{!}{%
\begin{tabular}{l l r r r r r r}
\toprule
Model & Treatment & MATH-500 & LCB & MBPP+ & SciKE & Overall & Cost \\
\midrule
\multirow{3}{*}{Qwen3-8B} & No compression & 86.87 / 96.80 & 28.88 / 46.56 & -- & -- & -- & -- \\
 & Tail (\ours{}) & 87.00 / 97.00 & 29.29 / 47.33 & 74.94 / 85.71 & 80.88 / 92.75 & 68.03 / 80.70 & 1.53 / 1,045 \\
 & Head and tail & 87.09 / 96.40 & 30.22 / 48.09 & 74.26 / 86.51 & 80.76 / 93.25 & 68.08 / 81.06 & 1.45 / 1,043 \\
\midrule
\multirow{5}{*}{Olmo-3-7B-Instruct} & LLM summary & -- & 31.04 / 56.49 & -- & 75.79 / 98.00 & -- & -- \\
 & Tail (\ours{}) & 93.11 / 97.40 & 32.63 / 58.78 & 76.17 / 87.04 & 75.82 / 97.75 & 69.43 / 85.24 & 6.26 / 1,235 \\
 & Head and tail & 93.52 / 97.80 & 32.35 / 56.49 & 75.29 / 86.77 & 76.66 / 98.25 & 69.45 / 84.83 & 6.39 / 1,247 \\
 & Summary block & -- & 24.52 / 47.33 & -- & 67.75 / 97.00 & -- & -- \\
 & Lowest log-prob & -- & 28.72 / 51.91 & -- & 76.27 / 97.50 & -- & -- \\
\bottomrule
\end{tabular}}
\caption{History compression ablation during exploration (\%). Each benchmark cell lists the mean accuracy over the 24 steps and the cumulative pass rate at step 24. Cost lists total GPU hours and PFLOPs over the four benchmarks. Dashes mark benchmarks that were not run. Single run with run seed 1.}
\label{tab:abl-compression}
\end{table*}

Removing the last response lowers the mean accuracy on every benchmark for both models. The largest drop is on LCB, about 8 pp on Qwen3-8B and 10 pp on Olmo-3-7B-Instruct. Its final pass rate on MATH-500 is slightly higher than that of \ours{} for both models.

\begin{table*}[t]
\centering
\small
\setlength{\tabcolsep}{3pt}
\resizebox{\textwidth}{!}{%
\begin{tabular}{l l r r r r r}
\toprule
Model & Update interval & MATH-500 & LCB & MBPP+ & SciKE & Overall / Total \\
\midrule
\multicolumn{7}{l}{\textit{Mean accuracy / cumulative pass rate (\%)}} \\
\midrule
\multirow{4}{*}{Qwen3-8B} & No update & 87.12 / 97.20 & 29.48 / 47.33 & 74.17 / 86.24 & 80.52 / 93.00 & 67.82 / 80.94 \\
 & Every step & 87.49 / 96.60 & 28.98 / 45.04 & 74.61 / 86.77 & 80.97 / 93.50 & 68.01 / 80.48 \\
 & Every 2 steps & 87.28 / 96.60 & 29.99 / 46.56 & 74.86 / 87.04 & 80.77 / 93.00 & 68.23 / 80.80 \\
 & Every 5 steps & 87.07 / 96.60 & 28.31 / 45.04 & 74.64 / 86.24 & 80.44 / 94.25 & 67.61 / 80.53 \\
\midrule
\multirow{4}{*}{Olmo-3-7B-Instruct} & No update & 93.68 / 97.40 & 32.38 / 55.73 & 76.17 / 87.57 & 76.13 / 97.25 & 69.59 / 84.49 \\
 & Every step & 93.59 / 97.60 & 31.96 / 56.49 & 76.36 / 87.30 & 76.10 / 99.25 & 69.52 / 85.16 \\
 & Every 2 steps & 93.27 / 97.60 & 31.97 / 54.20 & 76.18 / 86.77 & 75.94 / 98.00 & 69.34 / 84.14 \\
 & Every 5 steps & 93.57 / 98.00 & 32.00 / 52.67 & 76.23 / 87.57 & 76.72 / 98.75 & 69.63 / 84.25 \\
\midrule
\multicolumn{7}{l}{\textit{GPU h / PFLOPs / generated tokens (M)}} \\
\midrule
\multirow{4}{*}{Qwen3-8B} & No update & 0.48 / 350 / 21.4 & 0.46 / 140 / 8.6 & 0.22 / 173 / 10.6 & 0.32 / 380 / 23.2 & 1.47 / 1,044 / 63.7 \\
 & Every step & 2.07 / 1,507 / 31.9 & 1.31 / 312 / 10.2 & 0.98 / 739 / 15.8 & 1.19 / 999 / 29.1 & 5.55 / 3,557 / 86.9 \\
 & Every 2 steps & 1.57 / 930 / 26.6 & 1.24 / 226 / 9.4 & 0.70 / 454 / 13.1 & 0.86 / 693 / 26.2 & 4.38 / 2,302 / 75.3 \\
 & Every 5 steps & 1.03 / 542 / 23.1 & 1.09 / 170 / 8.8 & 0.61 / 266 / 11.4 & 0.68 / 484 / 24.2 & 3.40 / 1,463 / 67.5 \\
\midrule
\multirow{4}{*}{Olmo-3-7B-Instruct} & No update & 2.04 / 449 / 30.7 & 2.81 / 298 / 20.4 & 0.99 / 244 / 16.7 & 0.46 / 284 / 19.4 & 6.30 / 1,274 / 87.3 \\
 & Every step & 4.91 / 1,647 / 41.1 & 4.91 / 625 / 22.6 & 2.57 / 844 / 22.5 & 1.58 / 782 / 24.8 & 13.97 / 3,898 / 110.9 \\
 & Every 2 steps & 3.94 / 1,114 / 37.5 & 4.17 / 445 / 21.7 & 2.10 / 559 / 20.1 & 1.20 / 532 / 21.9 & 11.40 / 2,650 / 101.2 \\
 & Every 5 steps & 3.15 / 671 / 33.1 & 3.93 / 348 / 21.0 & 1.60 / 346 / 17.8 & 1.01 / 366 / 20.3 & 9.70 / 1,730 / 92.1 \\
\bottomrule
\end{tabular}}
\caption{Interleaved policy updates during exploration, by benchmark. The upper block lists the mean accuracy over the steps and the cumulative pass rate at the last step. The lower block lists the cumulative cost at the last step. The last column gives the Overall value for accuracy and the total over the four benchmarks for cost.}
\label{tab:abl-interleaved}
\end{table*}

\begin{table}[t]
\centering
\footnotesize
\setlength{\tabcolsep}{3pt}
\resizebox{\columnwidth}{!}{%
\begin{tabular}{l r r r r}
\toprule
Variant & Mean acc. & Final pass & GPU h & PFLOPs \\
\midrule
\ours{} & 68.705 & 82.715 & 3.885 & 1,159.0 \\
w/o last response & 63.395 & 81.095 & 3.620 & 926.0 \\
w/o feedback & 67.065 & 80.780 & 4.375 & 1,310.5 \\
Restart every 2 & 67.235 & 83.105 & 4.930 & 1,382.5 \\
Restart every 10 & 69.605 & 82.790 & 3.640 & 1,086.5 \\
No restart & 70.600 & 81.815 & 3.165 & 1,008.5 \\
Last 3 rounds & 69.285 & 82.670 & 4.290 & 1,520.0 \\
\bottomrule
\end{tabular}}
\caption{Component ablations averaged over Qwen3-8B and Olmo-3-7B-Instruct, underlying Figure~\ref{fig:ablation}. Mean acc.\ and Final pass are the averages of the Overall values in Table~\ref{tab:abl-components-acc} (\%). GPU h and PFLOPs are the averages of the totals in Table~\ref{tab:abl-components-cost}.}
\label{tab:abl-components-mean}
\end{table}

Removing the feedback lowers the LCB pass rate on Qwen3-8B by more than 8 pp and lowers the MBPP+ pass rate on both models. On Olmo-3-7B-Instruct SciKnowEval, w/o feedback has a higher mean accuracy than \ours{} and reaches a pass rate of 100\%.

Among the restart intervals, no restart gives the highest LCB mean accuracy on both models, but lower MBPP+ pass rates than \ours{} on both models. No restart also has the lowest total GPU hours on both models, and the last-3 variant has the highest total PFLOPs.

Table~\ref{tab:abl-components-mean} gives the two-model averages plotted in Figure~\ref{fig:ablation}. Restarting every 2 steps has the highest averaged final pass rate (83.105\%) and the highest GPU hours (4.930). No restart has the highest averaged mean accuracy (70.600\%) and the lowest GPU hours (3.165).

\begin{table*}[t]
\centering
\small
\setlength{\tabcolsep}{4.5pt}
\resizebox{\textwidth}{!}{%
\begin{tabular}{l l r r r r r r r}
\toprule
Model & Training set & Problems & Succ./problem & Fail./problem & Plain prompt & Random & From other problems & Restart \\
\midrule
\multirow{3}{*}{Qwen3-8B} & LCB & 307 & 12.93 & 6.96 & 69 & 3,393 & 1,118 & 578 \\
 & MATH & 432 & 15.96 & 4.02 & 217 & 5,567 & 3,574 & 1,326 \\
 & SciKnowEval & 334 & 11.13 & 3.78 & 165 & 2,671 & 1,698 & 1,046 \\
\midrule
\multirow{3}{*}{Olmo-3-7B-Instruct} & LCB & 334 & 11.59 & 8.40 & 33 & 3,223 & 558 & 648 \\
 & MATH & 469 & 18.32 & 1.63 & 328 & 6,932 & 5,307 & 1,658 \\
 & SciKnowEval & 375 & 11.68 & 8.13 & 46 & 3,467 & 872 & 912 \\
\bottomrule
\end{tabular}}
\caption{Trace statistics on the training problems. Problems is the number of problems with at least one successful context. Succ./problem and Fail./problem are the mean numbers of successful and failure contexts per problem. Plain prompt is the number of problems without a failure context. Random is the number of random-trace contexts with replaced feedback, and From other problems counts those whose feedback comes from another problem. Restart is the number of restart-step contexts kept unchanged.}
\label{tab:traces-stats}
\end{table*}

\paragraph{History compression.}
The tail length $L$ in \ours{} controls how a previous response longer than 2,000 tokens is shown in the memory. We compare the tail with five other ways of showing such a response:
\begin{itemize}
\item No compression shows the full response.
\item LLM summary shows the model's own summary of at most 1,000 tokens, generated as a further turn of the same conversation.
\item Head and tail shows the first 1,000 and the last 1,000 tokens.
\item Summary block asks every solving turn to end with a summary block. It always shows that block, cut to at most 500 tokens, followed by the final code or final answer.
\item Lowest log-prob shows the 1,000 tokens with the lowest sampled log-probability, in their original order.
\end{itemize}
These runs use an earlier version of \ours{} with different engine settings and chunked execution, so their GPU hours are not directly comparable with Table~\ref{tab:abl-components-cost}. No compression was run only on MATH-500 and LCB for Qwen3-8B. LLM summary, summary block and lowest log-prob were run only on LCB and SciKnowEval for Olmo-3-7B-Instruct. Table~\ref{tab:abl-compression} reports the results.

On Qwen3-8B, the tail and head-and-tail treatments are within 0.4 pp in Overall pass rate and close in cost. On Olmo-3-7B-Instruct, the tail has a higher Overall pass rate than head and tail. On Olmo-3-7B-Instruct LCB, the tail has the highest pass rate of all treatments, and the summary block has the lowest pass rate and the lowest mean accuracy. Against no compression on Qwen3-8B, the tail has slightly higher pass rates on MATH-500 and LCB.

\subsection{Interleaved updates}
\label{app:abl-interleaved}

\begin{table*}[t]
\centering
\small
\setlength{\tabcolsep}{3.2pt}
\begin{tabular}{l l r r r r r r r r r r}
\toprule
 & & \multicolumn{5}{c}{Qwen3-8B} & \multicolumn{5}{c}{Olmo-3-7B-Instruct} \\
\cmidrule(lr){3-7}\cmidrule(lr){8-12}
Set & Model & W$\to$R & R$\to$W & Repeat & Restart & Mean & W$\to$R & R$\to$W & Repeat & Restart & Mean \\
\midrule
\multirow{3}{*}{LCB} & Base & 26.26 & 3.59 & 7.71 & 45.60 & 65.80 & 29.73 & 16.15 & 0.17 & 48.20 & 57.81 \\
 & \ours{} & 14.35 & 2.72 & 8.43 & 57.49 & 65.60 & 30.27 & 12.87 & 0.10 & 55.54 & 63.77 \\
 & SFT & 19.17 & 10.71 & 9.00 & 47.72 & 55.60 & 25.26 & 13.19 & 0.44 & 49.55 & 58.20 \\
\midrule
\multirow{3}{*}{MATH} & Base & 15.92 & 2.37 & 64.80 & 76.50 & 79.86 & 24.92 & 1.10 & 47.54 & 89.66 & 92.24 \\
 & \ours{} & 25.84 & 1.91 & 43.82 & 91.44 & 92.29 & 26.04 & 1.15 & 39.93 & 89.98 & 92.60 \\
 & SFT & 12.66 & 3.12 & 57.12 & 62.73 & 67.73 & 23.67 & 0.97 & 56.42 & 84.01 & 88.76 \\
\midrule
\multirow{3}{*}{SciKE} & Base & 16.67 & 3.68 & 28.60 & 79.19 & 80.30 & 28.72 & 21.62 & 20.00 & 61.33 & 58.88 \\
 & \ours{} & 21.52 & 2.97 & 18.90 & 85.03 & 86.02 & 50.84 & 6.94 & 14.58 & 72.27 & 82.88 \\
 & SFT & 29.38 & 11.40 & 18.07 & 75.75 & 73.17 & 28.42 & 18.69 & 24.18 & 61.87 & 61.15 \\
\bottomrule
\end{tabular}
\caption{Self-correction on the training problems (\%). \ours{} is the policy trained with successful traces, and SFT is the SFT baseline. W$\to$R is wrong-to-right, R$\to$W is right-to-wrong, and Repeat is the repeated-error rate, all conditioned on the previous step. The wrong-to-wrong rate equals $100-$W$\to$R. Restart is the accuracy at restart steps, and Mean is the accuracy over all 10 steps. Single run with run seed 1.}
\label{tab:traces-selfcorr}
\end{table*}

This section reports the interleaved update results in Table~\ref{tab:interleaved} separately for each benchmark. The rollout runs at every step as in \ours{}. When the step index is a multiple of $N$, the policy update selects a batch, samples one student response per problem from the plain prompt and takes one optimizer step. The batch holds every problem with at least one correct, untruncated answer so far. For each problem, the teacher context is drawn uniformly from its successful contexts so far. The update uses the loss of Equation~\ref{eq:loss} and a learning rate of $10^{-6}$. Table~\ref{tab:abl-interleaved} reports accuracy, pass rate and cost for each benchmark, with \ours{} without updates as the first row of each block.

On Qwen3-8B, all three update intervals have lower LCB and MATH-500 pass rates than no update, and their SciKnowEval mean accuracies are within 0.5 percentage points of no update. On Olmo-3-7B-Instruct, all three update intervals have higher SciKnowEval pass rates than no update, with the highest from updating every step. Updating every step also has a higher Overall pass rate, measured after 22 steps on MATH-500 and LCB, while updating every 2 or 5 steps lowers the LCB pass rate. GPU hours increase on every benchmark for every update interval, by more than four times on Qwen3-8B MATH-500 when updating every step.

\subsection{Rollout traces in the policy update}
\label{app:abl-traces}

This section gives the construction and teacher accuracy of the three trace types in Table~\ref{tab:traces}, and the self-correction of the trained policy on the training problems. The three trace types differ only in the teacher context. The problems and the sampling settings are unchanged.
\begin{itemize}
\item Successful traces are the contexts in $\mathcal{C}^{+}(x)$.
\item Failure traces are the deduplicated contexts of the same problem whose following response is incorrect. A problem that is correct at all 20 rollout steps has no failure trace and uses the plain prompt instead.
\item Random traces replace the feedback turn of each non-restart successful context with a feedback text that differs from the original. The replacement comes from another step of the same problem when one exists and from another problem otherwise. On MATH and SciKnowEval, the feedback holds only the verdict, so this replacement flips the verdict. Contexts from restart steps hold no feedback and are kept unchanged.
\end{itemize}

\paragraph{Context statistics.}
Table~\ref{tab:traces-stats} lists the number of contexts of each type. Under failure traces, the share of problems that fall back to the plain prompt ranges from about 10\% on Olmo-3-7B-Instruct LCB to about 70\% on Olmo-3-7B-Instruct MATH. On MATH, most random traces take their feedback from another problem on both models.

\begin{table}[t]
\centering
\footnotesize
\setlength{\tabcolsep}{3pt}
\resizebox{\columnwidth}{!}{%
\begin{tabular}{l l r r r}
\toprule
Model & Traces & LCB & MATH & SciKE \\
\midrule
\multirow{4}{*}{Qwen3-8B} & Successful & 82.98 & 90.86 & 83.23 \\
 & Failure & 39.82 & 60.30 & 67.07 \\
 & \quad excl.\ plain prompt & 24.47 & 26.98 & 39.64 \\
 & Random & 69.30 & 88.54 & 80.16 \\
\midrule
\multirow{4}{*}{Olmo-3-7B-Inst.} & Successful & 71.03 & 96.16 & 67.93 \\
 & Failure & 44.76 & 79.48 & 50.13 \\
 & \quad excl.\ plain prompt & 39.87 & 46.99 & 44.22 \\
 & Random & 69.24 & 93.76 & 66.80 \\
\bottomrule
\end{tabular}}
\caption{Teacher accuracy (\%) under each trace type, extending Table~\ref{tab:traces}. The excl.\ plain prompt rows leave out problems without a failure context.}
\label{tab:traces-teacher}
\end{table}

\begin{table*}[t]
\centering
\small
\setlength{\tabcolsep}{3.5pt}
\begin{tabular}{l l r r r r}
\toprule
Model & Divergence & LCB v6 & MATH-500 & SciKE & Overall \\
\midrule
\multirow{4}{*}{Qwen3-8B} & Base & 19.08 / 31.30 & 84.45 / 93.60 & 74.69 / 80.50 & 59.41 \\
 & Reverse KL (\ours{}) & 26.15 / 37.40 & 91.75 / 96.60 & 76.88 / 82.25 & 64.92 \\
 & Forward KL & 23.66 / 37.40 & 91.75 / 95.80 & 76.00 / 85.25 & 63.80 \\
 & JSD & 24.81 / 33.59 & 78.65 / 89.20 & 75.62 / 83.50 & 59.69 \\
\midrule
\multirow{4}{*}{Olmo-3-7B-Instruct} & Base & 21.56 / 30.53 & 89.15 / 96.00 & 68.81 / 80.75 & 59.84 \\
 & Reverse KL (\ours{}) & 24.05 / 33.59 & 90.40 / 96.00 & 73.69 / 86.50 & 62.71 \\
 & Forward KL & 24.24 / 38.17 & 83.80 / 91.60 & 74.75 / 86.75 & 60.93 \\
 & JSD & 21.76 / 31.30 & 90.10 / 95.80 & 74.25 / 86.00 & 62.04 \\
\bottomrule
\end{tabular}
\caption{Divergence ablation by test set (\%). Each test cell lists avg@4 and pass@4. Overall is the mean avg@4 over LCB v6, MATH-500 and SciKnowEval, as in Table~\ref{tab:loss}. Single run with run seed 1.}
\label{tab:loss-bench}
\end{table*}

\paragraph{Teacher accuracy.}
The teacher accuracy in Table~\ref{tab:traces} is measured with the base model without LoRA. For each problem, the teacher receives four contexts. The context for sample $k$ is drawn as the trainer draws it at step $k$. The teacher generates one response per context with the test sampling settings, and the training verifier scores it. Table~\ref{tab:traces-teacher} also gives the failure-trace accuracy over only the problems that have at least one failure context. Excluding the plain-prompt problems lowers the failure-trace accuracy in every cell.

\paragraph{Self-correction on the training set.}
We also measure how the trained policies revise their own answers on the training problems. The \ours{} policy and the SFT baseline each run the exploration procedure of \ours{} for 10 steps on its training set, with restarts at steps 1 and 6. For the base model, we use the first 10 steps of the rollout that produced the policy-update data. We count only adjacent steps $(k,k+1)$ where $k+1$ is not a restart step, which gives 8 transitions per problem. We also count only the problems used in the policy update. Three rates are conditioned on the previous step:
\begin{itemize}
\item Wrong-to-right is the probability that step $k+1$ is correct, given that step $k$ is wrong.
\item Right-to-wrong is the probability that step $k+1$ is wrong, given that step $k$ is correct.
\item Repeat is the probability that step $k+1$ is wrong and has the same extracted final answer as step $k$, given that step $k$ is wrong.
\end{itemize}
Final answers are compared after removing whitespace. The extracted answer is the boxed answer on MATH, the program on LCB and the text after ``Final answer:'' on SciKnowEval. Table~\ref{tab:traces-selfcorr} reports these rates together with the restart-step accuracy and the mean accuracy. Both policies are the seed-1 checkpoints of Table~\ref{tab:general}.

On MATH and SciKnowEval, \ours{} has lower repeat rates than the base model on both models. On Olmo-3-7B-Instruct SciKnowEval, \ours{} also raises wrong-to-right from about 29\% for the base model to about 51\%, while SFT stays at the base-model level. On Qwen3-8B LCB, \ours{} instead has a lower wrong-to-right rate than the base model but a higher restart-step accuracy. SFT has a higher right-to-wrong rate than the base model on Qwen3-8B LCB and SciKnowEval, and a lower mean accuracy than the base model in four of the six cells.

\subsection{Divergence in the policy update}
\label{app:abl-loss}

This section reports the divergence ablation in Table~\ref{tab:loss} separately for each test set. As in Equation~\ref{eq:fdiv}, $p_i$ is the student distribution and $q_i$ is the teacher distribution. The three divergences are:
\begin{itemize}
\item reverse KL, which is $\mathrm{KL}(p_i\,\|\,q_i)$;
\item forward KL, which is $\mathrm{KL}(q_i\,\|\,p_i)$;
\item JSD with $\beta=0.5$, which is $0.5\,\mathrm{KL}(p_i\,\|\,m_i)+0.5\,\mathrm{KL}(q_i\,\|\,m_i)$ with $m_i=(p_i+q_i)/2$.
\end{itemize}
On MATH, each vocabulary term of every divergence is capped at $\tau=10^{-3}$. Only the divergence changes. The importance weights, token averaging, learning rate, teacher contexts, seed and engine settings are those of \ours{}, and all three variants use the same policy-update data. Table~\ref{tab:loss-bench} reports test performance, and Table~\ref{tab:loss-cost} reports truncation rates and training cost.

\begin{table}[t]
\centering
\footnotesize
\setlength{\tabcolsep}{2.5pt}
\resizebox{\columnwidth}{!}{%
\begin{tabular}{l l r r r r}
\toprule
Model & Div. & Trunc.\ (L / M / S) & GPU h (L / M / S) & GPU h & EFLOPs \\
\midrule
\multirow{4}{*}{Qwen3-8B} & Base & 0.00 / 0.05 / 0.00 & -- & -- & -- \\
 & RKL & 0.00 / 6.70 / 0.06 & 0.47 / 0.95 / 0.62 & 2.04 & 1.28 \\
 & FKL & 0.00 / 5.00 / 0.31 & 0.51 / 0.94 / 0.69 & 2.15 & 1.26 \\
 & JSD & 0.00 / 0.00 / 0.31 & 0.53 / 0.67 / 0.71 & 1.90 & 1.12 \\
\midrule
\multirow{4}{*}{Olmo-3-7B-I.} & Base & 2.67 / 3.25 / 0.19 & -- & -- & -- \\
 & RKL & 3.24 / 3.85 / 0.25 & 1.85 / 1.48 / 0.71 & 4.04 & 1.44 \\
 & FKL & 4.58 / 1.35 / 0.12 & 1.86 / 1.29 / 0.69 & 3.85 & 1.33 \\
 & JSD & 2.48 / 2.70 / 0.31 & 1.79 / 1.36 / 0.67 & 3.81 & 1.37 \\
\bottomrule
\end{tabular}}
\caption{Truncation rate of test responses (\%) and training cost of the divergence ablation. L / M / S denote LCB v6 / MATH-500 / SciKnowEval for truncation and the corresponding training sets for GPU hours. The mean of the three truncation rates gives the Trunc.\ column of Table~\ref{tab:loss}. Training cost excludes the data rollout.}
\label{tab:loss-cost}
\end{table}

\begin{table}[t]
\centering
\footnotesize
\setlength{\tabcolsep}{2.5pt}
\resizebox{\columnwidth}{!}{%
\begin{tabular}{l l r r r r r}
\toprule
Model & Set & \#Prob. & Base & Update & Rollout & Rel.\ gain \\
\midrule
\multirow{3}{*}{Qwen3-8B} & LCB & 307 & 42.35 & 58.31 & 72.64 & 52.69 \\
 & MATH & 432 & 73.15 & 86.34 & 81.71 & 154.05 \\
 & SciKE & 334 & 74.63 & 82.63 & 82.04 & 108.08 \\
\midrule
\multirow{3}{*}{Olmo-3-7B-I.} & LCB & 334 & 34.13 & 38.62 & 60.48 & 17.05 \\
 & MATH & 469 & 80.44 & 83.58 & 93.82 & 23.51 \\
 & SciKE & 375 & 44.47 & 70.67 & 61.07 & 157.83 \\
\bottomrule
\end{tabular}}
\caption{Internalization check on the training problems used in the policy update (\%). Base, Update and Rollout are the base model, policy update and with-context rollout accuracies. Rel.\ gain is defined in the text.}
\label{tab:internalization}
\end{table}

The MATH-500 drops stated in Section~\ref{sec:loss} are measured relative to reverse KL: 13.1 pp for JSD on Qwen3-8B and 6.6 pp for forward KL on Olmo-3-7B-Instruct. Both runs are also more than 5 pp below Base on MATH-500. On Qwen3-8B MATH-500, reverse KL has the highest truncation rate, followed by forward KL, and JSD has no truncated responses (Table~\ref{tab:loss-cost}). Forward KL has the highest SciKnowEval avg@4 on Olmo-3-7B-Instruct and the highest SciKnowEval pass@4 on Qwen3-8B. The training costs of the three divergences are about 2 GPU hours on Qwen3-8B and about 4 GPU hours on Olmo-3-7B-Instruct.

\paragraph{Internalization check.}
We test how much of the with-context rollout accuracy the policy keeps without the context. We fix the subset of training problems with at least one correct answer, which are the problems used in the policy update. On this subset, we compare three accuracies:
\begin{itemize}
\item Base acc.\ is the avg@4 of the base model under the test generation and judging settings. On LCB, it uses all tests of the training problem.
\item Policy-update acc.\ is the accuracy of the last student response for each problem during the policy update. This response is sampled from the plain prompt and judged offline in the same way.
\item Rollout acc.\ is the correctness rate of the step-20 rollout of \ours{} with context, which produced the policy-update data.
\end{itemize}
The relative gain is $(\text{Policy-update acc.}-\text{Base acc.})/(\text{Rollout acc.}-\text{Base acc.})$. A value above 100\% means that the policy without context exceeds the with-context rollout accuracy. Table~\ref{tab:internalization} reports the results.

The policy without context exceeds the with-context rollout accuracy in three cells: Qwen3-8B MATH, Qwen3-8B SciKnowEval and Olmo-3-7B-Instruct SciKnowEval. In the other three cells, the relative gain is about half on Qwen3-8B LCB and below a quarter on Olmo-3-7B-Instruct LCB and MATH. The policy update accuracy is above the base model accuracy in all six cells.

\end{document}